\documentclass{ecai}  

\usepackage{graphicx}
\usepackage{latexsym}
\usepackage{algorithm}
\usepackage{algorithmic}
\usepackage{amssymb,amsfonts,amsmath,amsthm}
\usepackage{multicol,multirow}
\usepackage{booktabs} 
\newtheorem{thm}{Theorem}
\newtheorem{remark}[thm]{Remark}
\newtheorem{lem}[thm]{Lemma}
\newtheorem{assumption}[thm]{Assumption}

\begin{document}

\begin{frontmatter}

\title{High Probability Analysis for Non-Convex Stochastic Optimization with Clipping}

\author[A;B]{\fnms{Shaojie}~\snm{Li}}
\author[A;B]{\fnms{Yong}~\snm{Liu} \thanks{Corresponding Author. Email: liuyonggsai@ruc.edu.cn.}}

\address[A]{Gaoling School of Artificial Intelligence, Renmin University of China, Beijing, China}
\address[B]{Beijing Key Laboratory of Big Data Management and Analysis Methods, Beijing, China}

\begin{abstract}
Gradient clipping is a commonly used technique to stabilize the training process of neural networks. A growing body of studies has shown that gradient clipping is a promising technique for dealing with the heavy-tailed behavior that emerged in stochastic optimization as well. While gradient clipping is significant, its theoretical guarantees are scarce. Most theoretical guarantees only provide an in-expectation analysis and only focus on optimization performance. In this paper, we provide high probability analysis in the non-convex setting and derive the optimization bound and the generalization bound simultaneously for popular stochastic optimization algorithms with gradient clipping, including stochastic gradient descent and its variants of momentum and adaptive stepsizes. With the gradient clipping, we study a heavy-tailed assumption that the gradients only have bounded $\alpha$-th moments for some $\alpha \in (1, 2]$, which is much weaker than the standard bounded second-moment assumption. Overall, our study provides a relatively complete picture for the theoretical guarantee of stochastic optimization algorithms with clipping.
\end{abstract}

\end{frontmatter}

\section{Introduction}
Stochastic optimization has played a crucial role in modern machine learning and data-driven optimization since many machine learning problems can be transformed into a stochastic optimization problem \cite{bottou201113,bottou2018optimization}. The past decades have witnessed the prosperous development of stochastic optimization algorithms. For example, stochastic gradient descent (SGD) \cite{robbins1951stochastic} has shown great success in the training of a large number of learning tasks \cite{jain2017non,hazan2016introduction}. In practice, SGD works by querying an oracle iteratively to obtain unbiased gradient estimates built on one or several training examples in place of the exact gradient. Its simplicity in implementation and low memory requirements per iteration make it easier to scale into the big data era \cite{lan2020first,bottou2010large}. 

Driven by the empirical success of SGD, a great deal of work has been done on design modifications to improve its performance in various ways. One popular modification is to use the adaptive stepsizes. \cite{duchi2011adaptive,mcmahan2010adaptive} propose the provably convergent adaptive gradient (AdaGrad) and demonstrate that the sparsity of the gradient suggests outperformance. Another popular modification of SGD is the momentum technique. Momentum uses a running average of the past gradient values \cite{polyak1964some,nesterov1983method}, and intuitively, adding momentum accelerates convergence by circumventing sharp curvatures and long ravines of the sub-level sets of the objective function \cite{ramezani2021generalization}. These stochastic optimization algorithms have shown distinct advantages in different learning tasks \cite{zhou2018convergence,zhang2020adaptive}. The superior empirical performance has attracted many researchers to investigate their guarantees and understand their theoretical properties.

Recently, a number of works have interestingly shown that stochastic optimization algorithms easily exhibit a heavy-tailed behavior \cite{zhang2020adaptive,simsekli2019tail,panigrahi2019non,csimcsekli2019heavy,camuto2021asymmetric,gurbuzbalaban2021heavy}. For example, \cite{zhang2020adaptive} provide empirical study and show that large natural language processing models, e.g., Bert \cite{vaswani2017attention,devlin2018bert}, have heavy-tailed gradients. In this spirit, existing guarantees of assuming bounded variance or light sub-Gaussian tail seem to be inappropriate \cite{zhang2020adaptive,zhang2019adam,cutkosky2021high}. In particular, in practice the variance can be very large, possibly even infinite, but the $\alpha$-th moment is bounded for some $\alpha \in (1, 2]$ \cite{cutkosky2021high,zhang2019adam}. For a more realistic analysis, it is essential to investigate the theoretical guarantees of stochastic optimization algorithms under this heavy-tailed condition. However, the setup becomes complicated, which hinders the use of conventional convergence analysis techniques that rely on the existence of the second-order moment.

Gradient clipping is an effective tool for dealing with heavy-tailed random variables \cite{cutkosky2021high,zhang2019adam}. The intuition behind this is that the clipped version of a heavy-tailed random variable will have much more benign properties when the clipping parameters are well chosen. Thus, gradient clipping is a promising technique for dealing with the heavy-tailed behavior in stochastic optimization. Additionally, gradient clipping can stabilize the gradient updates and thus stabilize the training process of stochastic optimization \cite{mai2021stability}. It is believed to effectively alleviate the gradient explosion problem without adding additional cost to the original update \cite{zhang2020improved}. As such, it has been a common choice for many application domains of machine learning, especially the language processing tasks \cite{pascanu2013difficulty,you2017scaling}.

Theoretically, some recent works have studied the optimization guarantee for stochastic optimization algorithms with gradient clipping \cite{gorbunov2020stochastic,zhang2020improved,mai2021stability,zhang2019gradient,zhang2019adam,zhang2020adaptive,cutkosky2021high}. However, these optimization guarantees are typically either provided for the convex optimization problems \cite{gorbunov2020stochastic,mai2021stability} or derived in expectation \cite{zhang2020improved,mai2021stability,zhang2019gradient,zhang2019adam,zhang2020adaptive}. Unfortunately, the expectation bound does not capture the behavior of stochastic optimization algorithms within one or several runs, which is relevant to the probabilistic property of stochastic optimization algorithms. Also, in real-world applications such as neural networks, since the training process can take hours or even days, algorithms are usually run only once, so it is important to obtain high probability guarantees \cite{li2020high,harvey2019tight,ward2019adagrad,cutkosky2021high,gorbunov2020stochastic}.

Furthermore, to the best of our knowledge, existing learning guarantees of stochastic optimization algorithms with clipping are almost all derived from the optimization performance perspective. In machine learning, our primary interest would be the generalization performance of the trained model on testing examples, which is quite different from the empirical performance on training examples \cite{lei2021learning,neyshabur2017exploring,bottou201113}. To be specific, the optimization performance concerns how the learning algorithm minimizes the empirical risk, while generalization performance concerns how the predictive models learned from training samples behave on the testing samples. Thus, to investigate the learning guarantees of clipped stochastic optimization algorithms, it is necessary to consider both the optimization and generalization guarantees.

Motivated by the problems we discussed above, this paper considers three popular stochastic optimization algorithms, i.e., stochastic gradient descent (SGD), stochastic gradient descent with momentum (SGDM), and stochastic gradient descent with adaptive stepsizes (SGDAS), in the non-convex setting. We establish both the high probability optimization bound and the high probability generalization bound for their clipped version under the bounded $\alpha$-th moment assumption. The results cover SGD and the well-known momentum technique and adaptive stepsizes and reveal the learning performance of the clipped stochastic optimization algorithms from both the perspective of convergence and generalization. In Table \ref{sample-table}, we provide an intuitive display of the results this paper obtained.

This paper is organized as follows. We first review the related work in Section \ref{related} and then introduce the preliminaries relevant to our discussion in Section \ref{oqwdrp;i}. Section \ref{mainresult} presents the main results, where we derive a series of learning guarantees for stochastic optimization algorithms with clipping. In Section \ref{concousion}, we conclude this paper. Some Lemmas useful to our discussions and proofs are shown in Section \ref{seclemma}, and the complete proofs are provided in the Appendix. 

\section{Related Work}\label{related}
\textbf{High Probability Bounds.}
Most of the literature provides guarantees in expectation for stochastic optimization algorithms \cite{harvey2019tight}. The high probability guarantees of SGD are mainly provided for the convex setting \cite{kakade2009generalization,hazan2014beyond,rakhlin2012making,gorbunov2020stochastic,harvey2019tight,davis2020high,davis2021low,gorbunov2021near,jain2019making,lei2018stochastic,london2017pac,lei2021generalization,feldman2019high,bassily2020stability}. As a comparison, high probability studies on the non-convex setting are scarce. Specifically, \cite{ghadimi2013stochastic,lei2021learning,madden2020high,li2022high} provide high probability bounds for non-convex SGD and \cite{li2020high,zhou2018convergence,ward2019adagrad,kavis2021high} for non-convex adaptive SGD. Unsatisfied, all these works assume the light sub-Gaussian tail or bounded variance. Very recently, motivated by recent research on the heavy-tailed phenomena in stochastic optimization, \cite{cutkosky2021high} give high probability bounds in the non-convex setting by assuming the bounded $\alpha$-th moment, a heavy-tailed assumption allowing unbounded variance. We mention to readers here that in some literature \cite{madden2020high,gorbunov2020stochastic,li2022high}, the “heavy-tailedness” refers to non-sub-Gaussianity. While in this paper, by stochastic gradient with heavy-tailed distribution, we mean such a stochastic gradient allows unbounded variance. Overall, high probability bounds for stochastic optimization algorithms under the heavy-tailed assumption allowing unbounded variance are scarce.

\noindent\textbf{Gradient Clipping.}
Gradient clipping is a commonly used technique in the training process of neural networks \cite{goodfellow2016deep,menon2019can}. In \cite{zhang2020improved,gorbunov2020stochastic,mai2021stability,zhang2019gradient,zhang2019adam,zhang2020adaptive,cutkosky2021high,li2022high}, the optimization guarantees of clipping are investigated. Specifically, \cite{gorbunov2020stochastic} study convex SGD and consider the smoothness and bounded variance conditions. \cite{mai2021stability} then study convex SGD but the non-smooth case. \cite{zhang2019gradient} study non-convex SGD, using a relaxed smoothness condition and a stronger assumption than the bounded variance. \cite{zhang2020improved} then provide improved convergence analysis of \cite{zhang2019gradient} with joint consideration of clipped gradient and clipped momentum. \cite{zhang2019adam,zhang2020adaptive} study non-convex SGD under the bounded $\alpha$-th moment condition. Notably that the above works all focus on in-expectation optimization guarantees. Under the bounded $\alpha$-th moment condition, \cite{cutkosky2021high} combine the gradient clipping and normalized gradient descent and derive the first high probability optimization guarantees for SGDM. The work \cite{li2022high} then provide the first high probability optimization guarantee for SGD with the sub-Weibull gradient noise. Therefore, from the related work, one can see that the high probability optimization analysis of non-convex stochastic optimization algorithms with clipping has not been thoroughly studied and is far from being understood. Even worse, there is almost no research on its generalization performance analysis. This paper makes an effort in this direction.
\section{Preliminaries}\label{oqwdrp;i}
\subsection{Notations}
Let $P$ be a probability measure defined on a sample space $\mathcal{Z}$,  many 
learning problems of machine learning  can be cast into  the following stochastic optimization problem  with a hypothesis space indexed by $\mathcal{W} \subseteq \mathbb{R}^d$:
\begin{align*}
\min_{\mathbf{w} \in \mathcal{W}} F(\mathbf{w}) := \mathbb{E}_{z \sim P} [f(\mathbf{w}; z)],
\end{align*}
  where the objective $f: \mathcal{W} \times \mathcal{Z} \mapsto \mathbb{R}_+$ is possibly non-convex and $\mathbb{E}_{z \sim P}$ denotes the expectation with respect to (w.r.t.) the random variable $z$ drawn form $P$. In machine learning, $F(\mathbf{w})$ is typically referred to as population risk \cite{bousquet2002stability}.

For the above stochastic optimization problem, people want to learn a prediction model with a small population risk. However, $F(\mathbf{w})$ is typically not accessible since the underlying distribution $P$ is unknown. In practice, we often sample a set of i.i.d. training data $S =  \{z_1,...,z_n \}$ from $P$ and minimize the following empirical risk: 
\begin{align*}
F_S(\mathbf{w}) := \frac{1}{n} \sum_{i=1}^n f(\mathbf{w}; z_i).
\end{align*}
Various stochastic optimization algorithms, e.g. SGD and its variants of momentum and adaptive stepsizes, have been proposed to optimize the empirical risk $F_S(\mathbf{w})$ and have shown their distinct advantages in different learning tasks \cite{bottou2018optimization,ghadimi2013stochastic}. Perhaps SGD is the most popular stochastic optimization algorithm due to its simplicity in implementation, low computational complexity, and sound practical behavior. For this reason, we show the pseudocode of SGD in Algorithm \ref{alg:exfeampletrht}. SGD iteratively moves models along the reverse
direction of an unbiased gradient estimate $\nabla f(\mathbf{w}_t;z_{j_t})$, i.e.,
\begin{align*}
\mathbb{E}_{j_t}[\nabla f(\mathbf{w}_t;z_{j_t}) - \nabla F_S(\mathbf{w}_{t})] = 0,
\end{align*}
and the simplicity has made SGD become one of the workhorses behind many machine learning tasks \cite{bottou201113,bottou2018optimization,lan2020first,lei2021learning}. Its clipped version and other variants will be presented in Section \ref{mainresult}.

We then introduce some notations used in this paper.
Let $b = \sup_{z \in \mathcal{Z}}\| \nabla f(0;z)\|$, where $\nabla f(\cdot;z)$ denotes the gradient of $f$ w.r.t. the first argument and $\| \cdot \|$ denotes the Euclidean norm. Let $B(\mathbf{0}, R) := \{ \mathbf{w} \in \mathbb{R}^d: \|\mathbf{w}-\mathbf{0}\| \leq R \}$ denote a ball with center $\mathbf{0}\in \mathbb{R}^d$ and radius $R$, denoted by $B_R$.
 We also denote $A \asymp B $ if there exists universal constants $C_1, C_2>0$  such that $C_1A \leq B \leq C_2 A$. Standard
order of magnitude notation such as $\mathcal{O}(\cdot)$ will be
used.
\begin{algorithm}[tb]
   \caption{SGD }
   \label{alg:exfeampletrht}
   \textbf{Input:} initial point $\mathbf{w}_1 = 0$, step sizes $\{ \eta_t \}_t$, dataset $S = \{z_1,...,z_n \}$.\\
        \vspace{-0.4cm}
\begin{algorithmic}[1]
    \FOR{$t=1,...,T$}
    \STATE  draw  $j_t$ from the uniform distribution over the set $\{ j: j\in [n] \}$ \\
   \STATE update $\mathbf{w}_{t+1} = \mathbf{w}_{t} - \eta_t \nabla f(\mathbf{w}_t;z_{j_t})$.
   \ENDFOR
\end{algorithmic}
\end{algorithm}

\subsection{Assumptions}
We first present the assumption of smoothness.
\begin{assumption}\label{ass1}
Let the constant $L > 0$. A differentiable function $g: \mathcal{W} \mapsto \mathbb{R}$ is $L$-smooth if
\begin{align*}
\| \nabla g(\mathbf{w}) -  \nabla g(\mathbf{w}')\| \leq  L \| \mathbf{w} -  \mathbf{w}' \|, \quad \forall \mathbf{w}, \mathbf{w}' \in \mathcal{W}.
\end{align*}
\end{assumption}
\begin{remark}\rm{}
This assumption is necessary to have the convergence of the gradients to zero \cite{li2020high}. It is standard in the optimization and generalization literature, e.g. \cite{feldman2019high,hardt2016train,reddi2016stochastic,foster2018uniform,cutkosky2021high,jain2017non}, to mention but a few. In this paper, for the optimization guarantees, we just need the empirical risk $F_S$ to be smooth, i.e., for any $\mathbf{w},\mathbf{w}' \in \mathcal{W}$, there holds $\| \nabla F_S(\mathbf{w}) -  \nabla F_S(\mathbf{w}')\| \leq L \| \mathbf{w} -  \mathbf{w}' \|$. While for the generalization guarantees, we need the function $f$ to be smooth, i.e., for any sample $z \in \mathcal{Z}$ and $\mathbf{w},\mathbf{w}' \in \mathcal{W}$, there holds $\| \nabla f(\mathbf{w};z) -  \nabla f(\mathbf{w}';z)\| \leq L \| \mathbf{w} -  \mathbf{w}' \|$.
\end{remark}
With the smoothness assumption, we have the useful ``descent lemma'' \cite{nesterov2014introductory}:
\begin{align*}
g(\mathbf{w}) - g(\mathbf{w}') \leq \langle \mathbf{w} -\mathbf{w}', \nabla g(\mathbf{w}') \rangle + \frac{L}{2} \| \mathbf{w} -\mathbf{w}'  \|^2.
\end{align*}
We then show our assumption on the stochastic gradient.
\begin{assumption}\label{brhbr}
There exists positive real numbers $\alpha \in (1, 2]$ and $G > 0$ such that for all $\mathbf{w}_t$,
\begin{align*}
\mathbb{E}_{j_t}[\|\nabla f(\mathbf{w}_t;z_{j_t})\|^{\alpha}] \leq  G^{\alpha}.
\end{align*}
\end{assumption}
\begin{remark}\rm{}
It is possible that the variance of $\nabla f(\mathbf{w}_t;z_{j_t})$ is unbounded while simultaneously satisfying Assumption \ref{brhbr} for $\alpha < 2$, e.g. the Pareto or $\alpha$-stable Levy random variables, please refer to Section 2.1 in \cite{zhang2019adam} for details. This assumption is thus much weaker than the standard bounded second moment assumption. It is shown that the unbounded variance strongly corrupts the optimization process and that previous convergence proofs for SGD fail  \cite{zhang2019adam}. Thus, it is essential to investigate the theoretical guarantees of stochastic optimization algorithms under this heavy-tailed condition.  This paper uses gradient clipping to establish high probability guarantees for many popular stochastic optimization algorithms under this assumption. 
\end{remark}

\section{Main Results}\label{mainresult}
In this section, we present the main results of this paper. We first consider SGD with gradient clipping in Section \ref{section3222}, and then SGDM with joint consideration of gradient clipping and momentum clipping in Section \ref{312}. Further, we study AdaGrad with gradient clipping in Section \ref{section431} and study a more general template of adaptive algorithms in Section \ref{section34}. 

In the general nonconvex case, since obtaining the global minimum is NP-hard in general, we cannot guarantee that the algorithm can find a global minimizer. Therefore, we are interested in finding the $\epsilon$-stationary point of first-order gradient for both the optimization guarantees and the generalization guarantees \cite{ghadimi2013stochastic,lei2021generalization,li2020high,madden2020high,zhang2020adaptive,cutkosky2021high}. 
\subsection{SGD with Clipping}\label{section3222}
\begin{algorithm}[tb]
   \caption{SGD with Clipping}
   \label{alg:gdcil}
   \textbf{Input:} initial point $\mathbf{w}_1 = 0$, step sizes $\{ \eta_t \}_t$, dataset $S = \{z_1,...,z_n \}$, and clipping parameter $\tau > 0$.\\
       \vspace{-0.4cm}
\begin{algorithmic}[1]
    \FOR{$t=1,...,T$}
    \STATE  draw  $j_t$ from the uniform distribution over the set $\{ j: j\in [n] \}$ \\
   \STATE obtain $\nabla \bar{f}(\mathbf{w}_t;z_{j_t})  = \frac{\nabla f(\mathbf{w}_t;z_{j_t})}{\|\nabla f(\mathbf{w}_t;z_{j_t})\|}\min \{ \tau, \|\nabla f(\mathbf{w}_t;z_{j_t})\| \}$ \\
   \STATE update $\mathbf{w}_{t+1} = \mathbf{w}_{t} - \eta_t \nabla \bar{f}(\mathbf{w}_t;z_{j_t})$.
   \ENDFOR
\end{algorithmic}
\end{algorithm}
The pseudocode of SGD with clipping is shown in Algorithm \ref{alg:gdcil}. In each iterate, SGD moves models along the reverse direction of a clipped gradient $\nabla \bar{f}(\mathbf{w}_t;z_{j_t})$, which is a biased estimate. We first present the optimization guarantee and then the generalization guarantee for clipped SGD.
\begin{thm}\label{theroem1}
Suppose the empirical risk $F_S$ satisfies Assumption \ref{ass1} and suppose Assumption \ref{brhbr} holds. Let $\mathbf{w}_t$ be the iterate produced by Algorithm \ref{alg:gdcil}. Set $\eta_t=\eta= p\frac{1}{T^{\frac{\alpha}{3\alpha-2}}}$ and $\tau = q T^{\frac{1}{3\alpha - 2}}$ for some positive constants $p,q$ such that $q\leq T^{\frac{2\alpha-2}{\alpha(3\alpha-2)}}$ and $\eta \leq 1/(12L)$.
Then for any $\delta \in (0,1)$, with probability $1-\delta$, we have
\begin{align*}
\frac{1}{T}\sum_{t=1}^T \| \nabla F_S(\mathbf{w}_{t}) \|^2
= \mathcal{O}\left(\frac{1}{T^\frac{2\alpha -2}{3\alpha - 2}}\log \frac{1}{\delta}\right).
\end{align*}
\end{thm}
\begin{remark}\rm{}
Theorem \ref{theroem1} suggests that if the empirical risk is smooth and the stochastic gradient follows from the heavy-tailed assumption, the optimization guarantee of clipped SGD has a convergence rate of the order $\mathcal{O}(\log (\frac{1}{\delta})/T^\frac{2\alpha -2}{3\alpha - 2})$. When $\alpha = 2$, it implies $\mathcal{O}(\log (\frac{1}{\delta})/T^\frac{1}{2})$. We now compare Theorem \ref{theroem1} with the related work of clipping. Theorem 3.1 in \cite{gorbunov2020stochastic} provides a high probability convergence bound for clipped SGD under the smoothness, convexity, and bounded variance conditions. Theorem 8 in \cite{zhang2019gradient} provides an in-expectation analysis for non-convex clipped SGD under a relaxed smoothness condition and a stronger assumption than the bounded variance, i.e., $\|\nabla f(\mathbf{w}_t;z_{j_t})- \nabla F_S(\mathbf{w}_{t})\| \leq G$ holds for any $\mathbf{w}_t$ and $z_{j_t}$ almost surely. The most relevant result to Theorem \ref{theroem1} is Theorem 2 in \cite{zhang2019adam}. Theorem \ref{theroem1} provides a high-probability result for non-convex clipped SGD, matching the in-expectation convergence rate of Theorem 2 in \cite{zhang2019adam} up to logarithmic factors. It has been shown  in \cite{cutkosky2021high} and Theorem 6 of \cite{zhang2020adaptive} that this rate is optimal. The benefit of the high probability bound is that it holds for any training data $S$ drawn from $P$ and over the randomness of the algorithm. To our best knowledge, Theorem \ref{theroem1} provides the first high probability optimization bound for clipped SGD under an unbounded variance assumption. We sketch the proof technique of Theorem \ref{theroem1}. The proof begins with the ``descent lemma'' and some decompositions, resulting in Eq. (1) in the Appendix. Unlike the in-expectation analysis, the high probability analysis requires to construct some martingale difference sequences, e.g. $\sum_{t=1}^T  L \eta^2 (\|\nabla \bar{f}(\mathbf{w}_t;z_{j_t}) - \mathbb{E}_{j_t}\nabla \bar{f}(\mathbf{w}_t;z_{j_t})\|^2-\mathbb{E}_{j_t}\|\nabla \bar{f}(\mathbf{w}_t;z_{j_t}) - \mathbb{E}_{j_t}\nabla \bar{f}(\mathbf{w}_t;z_{j_t})\|^2)$ and $- \sum_{t=1}^T  \eta    \langle \nabla \bar{f}(\mathbf{w}_t;z_{j_t}) - \mathbb{E}_{j_t}\nabla \bar{f}(\mathbf{w}_t;z_{j_t}), \nabla F_S(\mathbf{w}_{t}) \rangle$.  Some concentration inequalities on martingales should be used to bound these terms. The key point lies in that the Auzan-Hoeffding inequality for martingales with bounded increments fails to give the optimal rate of Theorem \ref{theroem1}, especially when dealing with $\sum_{t=1}^T  L \eta^2 (\|\nabla \bar{f}(\mathbf{w}_t;z_{j_t}) - \mathbb{E}_{j_t}\nabla \bar{f}(\mathbf{w}_t;z_{j_t})\|^2-\mathbb{E}_{j_t}\|\nabla \bar{f}(\mathbf{w}_t;z_{j_t}) - \mathbb{E}_{j_t}\nabla \bar{f}(\mathbf{w}_t;z_{j_t})\|^2)$. For the purpose of the optimal rate, one must consider the conditional variance and use the Bernstein-type concentration inequality (Lemma \ref{lemma5}). Notably that for clipped SGD, its conditional variance should be carefully controlled. Other terms like the bias $\sum_{t=1}^T L \eta^2\|\mathbb{E}_{j_t}\nabla \bar{f}(\mathbf{w}_t;z_{j_t}) -  \nabla F_S(\mathbf{w}_{t})\|^2$ and the variance $\sum_{t=1}^T L \eta^2 \mathbb{E}_{j_t}\|\nabla \bar{f}(\mathbf{w}_t;z_{j_t}) - \mathbb{E}_{j_t}\nabla \bar{f}(\mathbf{w}_t;z_{j_t})\|^2$ of the clipped stochastic gradient can be bounded by its boundedness and the heavy-tailed assumption. After getting the bound in Eq. (8) in the Appendix, carefully selecting the stepsize $\eta$ and clipping parameter $\tau$ obtains the optimal rate of Theorem \ref{theroem1}. 
\end{remark}
\begin{thm}\label{theorem22}
Suppose the function $f$ satisfies Assumption \ref{ass1} and suppose Assumption \ref{brhbr} holds. Let $\mathbf{w}_t$ be the iterate produced by Algorithm \ref{alg:gdcil}. Set $\eta_t=\eta= p\frac{1}{T^{\frac{\alpha}{3\alpha-2}}}$ and $\tau = q T^{\frac{1}{3\alpha - 2}}$ for some positive constants $p,q$ such that $q\leq T^{\frac{2\alpha-2}{\alpha(3\alpha-2)}}$ and $\eta \leq 1/(12L)$. Select $T \asymp (\frac{n}{d})^{\frac{3\alpha-2}{4\alpha-4}}$.
Then for any $\delta \in (0,1)$, with probability $1-\delta$, we have
\begin{align*}
\frac{1}{T}\sum_{t=1}^T \| \nabla F(\mathbf{w}_{t}) \|^2
= \mathcal{O}\left(\Big(\frac{d}{n}\Big)^{\frac{1}{2}}\log \frac{1}{\delta}\right).
\end{align*}
\end{thm}
\begin{remark}\rm{}\label{rmark22}
Theorem \ref{theorem22} shows that if the function $f$ is smooth and the stochastic gradient follows from the heavy-tailed assumption, the generalization guarantee of clipped SGD has a convergence rate of the order $\mathcal{O}((\frac{d}{n})^{\frac{1}{2}}\log \frac{1}{\delta})$ when the iterate number $T \asymp (\frac{n}{d})^{\frac{3\alpha-2}{4\alpha-4}}$. Lemma 4.3 in \cite{hardt2016train} provides an in-expectation analysis for clipped SGD by the lens of algorithmic stability \cite{bousquet2002stability}. For the generalization analysis of clipped stochastic optimization algorithms, we have not found other related results in the literature. The proof of Theorem \ref{theorem22} begins with a decomposition, resulting in Eq. (14) in the Appendix, where $\sum_{t=1}^T \|  \nabla F_S(\mathbf{w}_{t}) \|^2$ corresponds to Theorem \ref{theroem1} and  $T  \| \nabla F(\mathbf{w}_{T}) - \nabla F_S(\mathbf{w}_{T}) \|^2$ can be bounded by the uniform convergence of gradients (Lemma \ref{lemmab3}). In using Lemma \ref{lemmab3}, we need to quantify the value of $R_T$, which reveals the space complexity induced by the iterate update of SGD. In this spirit, we need to give the bound of SGD's iterate $\max_{1\leq t \leq T} \| \mathbf{w}_{t}\|$, see Eq. (11) in the Appendix. We show that this term can be bounded by the bias of the clipped stochastic gradient, the empirical risk, and the Pinelis-Bernstein inequality for martingales difference sequences (Lemma \ref{lemma51}). Again, the conditional variance should be carefully controlled to guarantee the convergence rate of Theorem \ref{theorem22} when using Lemma \ref{lemma51}. One can see that  $\frac{1}{T}\sum_{t=1}^T \| \nabla F_S(\mathbf{w}_{t}) \|^2$ is decreasing along the training process, while $\| \nabla F(\mathbf{w}_{T}) - \nabla F_S(\mathbf{w}_{T}) \|^2$ is increasing, which suggests the space complexity is keeping grow along the training process. Thus,  Theorem \ref{theorem22} reveals that an implicit regularization can be achieved by tuning the number of passes to balance the optimization and generalization error for the clipped stochastic gradient descent.
\end{remark}
\subsection{SGDM with Clipping}\label{312}
\begin{algorithm}[tb]
   \caption{SGDM with Clipping}
   \label{alg:thnntn}
   \textbf{Input:} initial point $\mathbf{w}_1 = 0$, $\mathbf{m}_0 = 0$, step sizes $\{ \eta_t \}_t$, dataset $S = \{z_1,...,z_n \}$, momentum parameter $\gamma$, and clipping parameters $\tau_1, \tau_2 > 0$.\\
       \vspace{-0.4cm}
\begin{algorithmic}[1]
    \FOR{$t=1,...,T$}
    \STATE  draw  $j_t$ from the uniform distribution over the set $\{ j: j\in [n] \}$ \\
   \STATE obtain $\nabla \bar{f}(\mathbf{w}_t;z_{j_t})  = \frac{\nabla f(\mathbf{w}_t;z_{j_t})}{\|\nabla f(\mathbf{w}_t;z_{j_t})\|}\min \{ \tau_1, \|\nabla f(\mathbf{w}_t;z_{j_t})\| \}$ \\
   \STATE update $\mathbf{m}_t = \gamma \mathbf{m}_{t-1} + (1 - \gamma) \nabla \bar{f}(\mathbf{w}_t;z_{j_t})$\\
   \STATE obtain $ \bar{\mathbf{m}}_t  = \frac{\mathbf{m}_t}{\|\mathbf{m}_t\|}\min \{ \tau_2, \|\mathbf{m}_t\| \}$ \\
   \STATE update $\mathbf{w}_{t+1} = \mathbf{w}_{t} - \eta_t \bar{\mathbf{m}}_t$.
   \ENDFOR
\end{algorithmic}
\end{algorithm}
The pseudocode of SGDM with clipping is shown in Algorithm \ref{alg:thnntn}. Algorithm \ref{alg:thnntn} incorporates the momentum update, $\mathbf{m}_t = \gamma \mathbf{m}_{t-1} + (1 - \gamma) \nabla \bar{f}(\mathbf{w}_t;z_{j_t})$, to SGD. We first give the optimization guarantee and then the generalization guarantee for clipped SGDM.
\begin{thm}\label{theorem33}
Suppose the empirical risk $F_S$ satisfies Assumption \ref{ass1} and suppose Assumption \ref{brhbr} holds. Let $\mathbf{w}_t$ be the iterate produced by Algorithm \ref{alg:thnntn}. Set $\tau_1=\frac{pG}{(1-\gamma)^{1/\alpha}}$, $1-\gamma = \frac{s}{T^{\frac{\alpha}{3\alpha-2}}}$, $\eta_t = \eta = \frac{q}{T^{\frac{\alpha }{3\alpha-2}}}$, and $\tau_2 = \frac{r}{T^{\frac{\alpha-1}{3\alpha-2}}}$ for some positive constants $p,s,q,r$ such that $1-\gamma \leq 1$.
Then for any $\delta \in (0,1)$, with probability $1-\delta$, we have
\begin{align*}
\frac{1}{T}\sum_{t=1}^T \| \nabla F_S(\mathbf{w}_{t}) \|
= \mathcal{O}\left(\frac{1}{T^\frac{\alpha -1}{3\alpha - 2}}\log \frac{T}{\delta}\right).
\end{align*}
\end{thm}
\begin{remark}\rm{}\label{remarks}
Theorem \ref{theorem33} shows that the optimization guarantee of SGDM with gradient clipping and momentum clipping has a convergence rate of the order $\mathcal{O}(\log \frac{T}{\delta}/T^\frac{\alpha -1}{3\alpha - 2})$. Note that according to Jensen's inequality, the bound in Theorem \ref{theroem1} implies that $\frac{1}{T}\sum_{t=1}^T \| \nabla F_S(\mathbf{w}_{t}) \|= ((\frac{1}{T}\sum_{t=1}^T \| \nabla F_S(\mathbf{w}_{t}) \|)^2)^{1/2} \leq (\frac{1}{T}\sum_{t=1}^T\| \nabla F_S(\mathbf{w}_{t}) \|^2)^{1/2} \leq \mathcal{O}(\log \frac{1}{\delta}/T^\frac{\alpha -1}{3\alpha - 2})$. Thus, Theorem \ref{theorem33} presents a similar order bound to Theorem \ref{theroem1}. An improvement of Theorem \ref{theorem33} is that its stepsize $\eta$ does not depend on the smoothness parameter $L$, i.e., completely oblivious to the knowledge of smoothness. We now compare Theorem \ref{theorem33} with the related work of clipping. As we discussed in Section \ref{related}, \cite{zhang2020improved} also study SGDM with both gradient clipping and momentum clipping. Their updates are $\mathbf{m}_{t+1} = \gamma \mathbf{m}_{t} + (1 - \gamma) \nabla f(\mathbf{w}_t;z_{j_t})$ and then $\mathbf{w}_{t+1} = \mathbf{w}_{t} - [v \min(\eta, \frac{\tau}{\|\mathbf{m}_{t+1}\|})\mathbf{m}_{t+1} +(1-v) \min(\eta, \frac{\tau}{\|\nabla f(\mathbf{w}_t;z_{j_t})\|})\nabla f(\mathbf{w}_t;z_{j_t})]$, where $v \in [0,1]$ is an interpolation parameter. Theorem 3.2 in \cite{zhang2020improved} provides an expected optimization bound under a relaxed smoothness and a stronger assumption than the bounded variance, i.e., $\|\nabla f(\mathbf{w}_t;z_{j_t})- \nabla F_S(\mathbf{w}_{t})\| \leq G$ holds for any $\mathbf{w}_t$ and $z_{j_t}$ almost surely, where the latter assumption is restrictive, hindering the scope of application of their results. Another related work is \cite{cutkosky2021high}. Theorem 2 in \cite{cutkosky2021high} gives a high-probability bound under the same conditions to Theorem \ref{theorem33} by combining the gradient clipping, momentum, and normalized momentum. Their updates are $\mathbf{m}_{t} = \gamma \mathbf{m}_{t-1} + (1 - \gamma) \nabla \bar{f}(\mathbf{w}_t;z_{j_t})$ and then $\mathbf{w}_{t+1} = \mathbf{w}_{t} - \eta_t \frac{\mathbf{m}_t}{\|\mathbf{m}_t\|}$. In Algorithm \ref{alg:thnntn}, we study the clipped version of momentum. Algorithm \ref{alg:thnntn} is more similar to the framework proposed in \cite{zhang2020improved}, where both the gradient clipping and momentum clipping are all considered. The proof techniques between ours and \cite{zhang2020improved,cutkosky2021high} are different. We now compare Theorem \ref{theorem33} with \cite{cutkosky2021high} considering the two works all focus on high probability bound. Due to $\|\frac{\mathbf{m}_t}{\|\mathbf{m}_t\|} \| = 1$, \cite{cutkosky2021high} show that $F_S(\mathbf{w}_{t+1}) - F_S(\mathbf{w}_{t}) \leq -\eta \langle  \frac{\mathbf{m}_t}{\|\mathbf{m}_t\|}, \nabla F_S(\mathbf{w}_{t}) \rangle + \frac{L}{2}\eta^2 =  -\eta \langle  \frac{\mathbf{m}_t}{\|\mathbf{m}_t\|} , \nabla F_S(\mathbf{w}_{t}) - \mathbf{m}_t \rangle - \eta \| \mathbf{m}_t  \| + \frac{L}{2}\eta^2 \leq  \eta \| \frac{\mathbf{m}_t}{\|\mathbf{m}_t\|} \| \|\nabla F_S(\mathbf{w}_{t}) - \mathbf{m}_t\| - \eta\| \mathbf{m}_t - \nabla F_S(\mathbf{w}_{t}) + \nabla F_S(\mathbf{w}_{t}) \| + \frac{L}{2}\eta^2 \leq 2\eta \|\nabla F_S(\mathbf{w}_{t}) - \mathbf{m}_t \| - \eta\|   \nabla F_S(\mathbf{w}_{t}) \| + \frac{L}{2}\eta^2$, which implies $\|   \nabla F_S(\mathbf{w}_{t}) \| \leq 2 \|\nabla F_S(\mathbf{w}_{t}) - \mathbf{m}_t \| + \frac{L}{2}\eta-\frac{(F_S(\mathbf{w}_{t+1}) - F_S(\mathbf{w}_{t}))}{\eta}$. \cite{cutkosky2021high} then use Freedman's inequality to bound the term $\|\nabla F_S(\mathbf{w}_{t}) - \mathbf{m}_t \|$. However, the clipped momentum doesn't have the property $\|\frac{\mathbf{m}_t}{\|\mathbf{m}_t\|} \| = 1$. In the proof of Theorem \ref{theorem33}, we need to consider two cases, i.e., $\| \mathbf{m}_t \| \geq \tau_2$ and $\| \mathbf{m}_t \| < \tau_2$. In the former case, we need to prove that $\| \nabla F_S(\mathbf{w}_{t})\| \leq  3\frac{F_S(\mathbf{w}_{t}) - F_S(\mathbf{w}_{t+1})}{\eta \tau_2}+4 \| \mathbf{m}_t - \nabla F_S(\mathbf{w}_{t}) \| + \frac{3L}{2}  \eta \tau_2$, and in the latter case, we need to prove that $\| \nabla F_S(\mathbf{w}_{t} ) \|^2 \leq   \frac{2(F_S(\mathbf{w}_{t}) - F_S(\mathbf{w}_{t+1}))}{\eta} + \| \nabla F_S(\mathbf{w}_{t}) - \mathbf{m}_t\|^2  +  L \eta \tau_2^2$. We then use the Pinelis-Bernstein inequality for martingales difference sequences (Lemma \ref{lemma51}) to bound the terms $\| \nabla F_S(\mathbf{w}_{t}) - \mathbf{m}_t\|$ and $\| \nabla F_S(\mathbf{w}_{t}) - \mathbf{m}_t\|^2$. Additionally, in practice, a more common application of the normalized momentum should be $\mathbf{w}_{t+1} = \mathbf{w}_{t} - \eta_t \frac{\mathbf{m}_t}{\|\mathbf{m}_t\| + \beta}$ with $\beta > 0$. However, this pattern of iterate update violates the property $\|\frac{\mathbf{m}_t}{\|\mathbf{m}_t\|} \| = 1$, which plays an essential role in the proof in \cite{cutkosky2021high}. It is unclear whether the proof techniques of \cite{cutkosky2021high} can guarantee the convergence for this more commonly used pattern of iterate update. The clear motivations of our study on momentum clipping include that the clipping doesn't have such an issue and is more common in practice, that Appendix C in \cite{zhang2020improved} suggests that there are some practical issues that make normalized momentum less favorable than traditional clipping methods, and that \cite{zhang2020improved} only provide in-expectation analysis. Considering the above analysis, we believe that Theorem \ref{theorem33} is an important result for stochastic optimization with clipping.
\end{remark}
\begin{thm}\label{theorem44}
Suppose the function $f$ satisfies Assumption \ref{ass1} and suppose Assumption \ref{brhbr} holds. Let $\mathbf{w}_t$ be the iterate produced by Algorithm \ref{alg:thnntn}. Set $\tau_1=\frac{pG}{(1-\gamma)^{1/\alpha}}$, $1-\gamma = \frac{s}{T^{\frac{\alpha}{3\alpha-2}}}$, $\eta_t = \eta = \frac{q}{T^{\frac{\alpha }{3\alpha-2}}}$, and $\tau_2 = \frac{r}{T^{\frac{\alpha-1}{3\alpha-2}}}$ for some positive constants $p,s,q,r$ such that $1-\gamma \leq 1$. Select $T \asymp (\frac{n}{d})^{\frac{3\alpha-2}{4\alpha-4}}$.
Then for any $\delta \in (0,1)$, with probability $1-\delta$, we have
\begin{align*}
\frac{1}{T}\sum_{t=1}^T \| \nabla F(\mathbf{w}_{t}) \|
= \mathcal{O}\left(\Big(\frac{d}{n}\Big)^{\frac{1}{4}}\log\frac{n}{d\delta}\right).
\end{align*}
\end{thm}
\begin{remark}\rm{}
According to Jensen's inequality, Theorem \ref{theorem44} shows a generalization bound of a similar order to Theorem \ref{theorem22}. For the generalization analysis of clipped SGDM and even SGDM, we have not found related results in the literature. The analysis pattern of Theorem \ref{theorem44} follows Remark \ref{rmark22}. Thus, Theorem \ref{theorem44} also reveals that an implicit regularization can be achieved by tuning the number of passes to balance the optimization and generalization error for the clipped stochastic gradient descent with momentum.
\end{remark}
\subsection{SGDAS with Clipping}\label{section3433}
After the momentum technique, this section studies SGD with the adaptive stepsizes. We first consider AdaGrad and then a more general form of adaptive accelerated algorithms, including AdaGrad and adaptive RSAG as specific examples.
\subsubsection{AdaGrad}\label{section431}
\begin{algorithm}[tb]
   \caption{AdaGrad with Clipping}
   \label{alg:thnntn3}
   \textbf{Input:} initial point $\mathbf{w}_1 = 0$, step sizes $\{ \eta_t \}_t$, dataset $S = \{z_1,...,z_n \}$, $G_0>0$, and $\tau > 0$.\\
       \vspace{-0.4cm}
\begin{algorithmic}[1]
    \FOR{$t=1,...,T$}
    \STATE  draw  $j_t$ from the uniform distribution over the set $\{ j: j\in [n] \}$ \\
   \STATE obtain $\nabla \bar{f}(\mathbf{w}_t;z_{j_t})  = \frac{\nabla f(\mathbf{w}_t;z_{j_t})}{\|\nabla f(\mathbf{w}_t;z_{j_t})\|}\min \{ \tau, \|\nabla f(\mathbf{w}_t;z_{j_t})\| \}$ \\
      \STATE obtain $\eta_t  = \frac{1}{\sqrt{G_0^2 + \sum_{k=1}^t \|\nabla \bar{f}(\mathbf{w}_t;z_{j_t})\|^2} }$ \\
   \STATE update $\mathbf{w}_{t+1} = \mathbf{w}_{t} - \eta_t \nabla \bar{f}(\mathbf{w}_t;z_{j_t})$.
   \ENDFOR
\end{algorithmic}
\end{algorithm}
The pseudocode of AdaGrad with clipping is shown in Algorithm \ref{alg:thnntn3}. Compared to the original AdaGrad \cite{duchi2011adaptive,mcmahan2010adaptive}, in each iterate, Algorithm \ref{alg:thnntn3} uses a clipped gradient estimate $\nabla \bar{f}(\mathbf{w}_t;z_{j_t})$. We first give the optimization guarantee and then the generalization guarantee.
\begin{thm}\label{thoremss}
Suppose the empirical risk $F_S$ satisfies Assumption \ref{ass1} and suppose Assumption \ref{brhbr} holds.  Assume that $F_S(\mathbf{w}) \leq M$ for all $\mathbf{w}$ for some $M$. Let $\mathbf{w}_t$ be the iterate produced by Algorithm \ref{alg:thnntn3}.  Set $\tau = p T^{\frac{1}{3\alpha - 2}}$ for some positive constants $p$ such that $p\leq T^{\frac{2\alpha-2}{\alpha(3\alpha-2)}}$.
Then for any $\delta \in (0,1)$, with probability $1-\delta$, we have
\begin{align*}
\frac{1}{T}\sum_{t=1}^T \| \nabla F_S(\mathbf{w}_{t}) \|^2
= \mathcal{O}\left(\frac{1}{T^\frac{2\alpha -2}{3\alpha - 2}}\log \frac{1}{\delta}\right).
\end{align*}
\end{thm}
\begin{remark}\rm{}
Theorem \ref{thoremss} shows that if $F_S$ is smooth and bounded and the stochastic gradient follows from the heavy-tailed assumption, the optimization guarantee of clipped AdaGrad has a convergence rate of the order $\mathcal{O}(\log \frac{1}{\delta}/T^\frac{2\alpha -2}{3\alpha - 2})$. Theorem \ref{thoremss} requires $F_S$ to be bounded additionally. This assumption also appears in Theorem 4 of \cite{cutkosky2020momentum} and Theorem 6 of \cite{tran2021better} when they prove the convergence rate for adaptive algorithms. To our best knowledge, Theorem \ref{thoremss} provides the first optimization bound for AdaGrad with clipping. The proof technique of clipped AdaGrad is different from the clipped SGD and SGDM. With the ``descent lemma'' and some decompositions, we instead prove that $\sum_{t=1}^T\| \nabla F_S(\mathbf{w}_{t}) \|^2 \leq (2M + L)\sqrt{G_0^2 + \sum_{t=1}^T \|\nabla \bar{f}(\mathbf{w}_t;z_{j_t})\|^2} - \sum_{t=1}^T \langle  \nabla \bar{f}(\mathbf{w}_t;z_{j_t}) - \nabla F_S(\mathbf{w}_{t}), \nabla F_S(\mathbf{w}_{t}) \rangle$ for clipped AdaGrad. Then, we need to bound the terms $\sum_{t=1}^T \|\nabla \bar{f}(\mathbf{w}_t;z_{j_t})\|^2$ and $- \sum_{t=1}^T \langle  \nabla \bar{f}(\mathbf{w}_t;z_{j_t}) - \nabla F_S(\mathbf{w}_{t}), \nabla F_S(\mathbf{w}_{t}) \rangle$ with the term $\sum_{t=1}^T\| \nabla F_S(\mathbf{w}_{t}) \|^2$, see Eqs. (21), (22) and (24) in the Appendix for details. To guarantee the optimal rate of Theorem \ref{thoremss}, in bounding the two terms, we need to use the Bernstein-type concentration inequality (Lemma \ref{lemma5}) since the Auzan-Hoeffding inequality for martingales with bounded increments leads to the sub-optimal rates. During this process, the conditional variance must be carefully considered. Finally, solving the quadratic inequality of $\sum_{t=1}^T \| \nabla F_S(\mathbf{w}_{t}) \|^2$, we get the optimization bound of Theorem \ref{thoremss}. We now compare Theorem \ref{thoremss} with the results of clipped SGD and clipped SGDM (Theorem \ref{theroem1} and Theorem \ref{theorem33}).  An improvement of Theorem \ref{thoremss} is that compared to clipped SGD, the stepsize $\eta_t$ of clipped AdaGrad does not depend on the smoothness parameter $L$ and the parameter $\alpha$ of Assumption \ref{brhbr} and compared to clipped SGDM, the stepsize $\eta_t$ of clipped AdaGrad does not depend on the parameter $\alpha$ of Assumption \ref{brhbr}.
\end{remark}

\begin{thm}\label{theorem789}
Suppose the function $f$ satisfies Assumption \ref{ass1} and suppose Assumption \ref{brhbr} holds. Assume that $F_S(\mathbf{w}) \leq M$ for all $\mathbf{w}$ for some $M$. Let $\mathbf{w}_t$ be the iterate produced by Algorithm \ref{alg:thnntn3}.  Set $\tau = p T^{\frac{1}{3\alpha - 2}}$ for some positive constants $p$ such that $p\leq T^{\frac{2\alpha-2}{\alpha(3\alpha-2)}}$. Select $T \asymp (\frac{n}{d})^{\frac{3\alpha-2}{4\alpha-4}}$.
Then for any $\delta \in (0,1)$, with probability $1-\delta$, we have 
\begin{align*}
\frac{1}{T}\sum_{t=1}^T \| \nabla F(\mathbf{w}_{t}) \|^2
=  \mathcal{O} \Big( \Big(\frac{d}{n}\Big)^{\frac{2\alpha-2}{5\alpha-4}} \log(\frac{1}{\delta}) \log \left(1+(\frac{n}{d})^{\frac{3\alpha}{5\alpha-4}}\right)\Big).
\end{align*}
\end{thm}
\begin{remark}\rm{}
Theorem \ref{theorem789} shows that 
the generalization guarantee of clipped AdaGrad has a convergence rate of the order $\mathcal{O}( (\frac{d}{n})^{\frac{2\alpha-2}{5\alpha-4}} \log(1/\delta) \log (1+(\frac{n}{d})^{\frac{3\alpha}{5\alpha-4}}))$ when the iterate number $T \asymp (\frac{n}{d})^{\frac{3\alpha-2}{4\alpha-4}}$. When $\alpha = 2$, Theorem \ref{theorem789} implies $\mathcal{O}( (\frac{d}{n})^{\frac{1}{3}} \log(1/\delta) \log (1+\frac{n}{d}))$. For the generalization analysis of clipped AdaGrad and even AdaGrad, we have not found related results in the literature. The analysis pattern of Theorem \ref{theorem789} follows Remark \ref{rmark22} and also reveals the implicit regularization effect. Investigating whether the generalization bound of clipped AdaGrad can achieve the similar order to SGD or SGDM is an interesting open problem. 
\end{remark}
\subsubsection{Adaptive Accelerated Algorithms}\label{section34}

\begin{algorithm}[tb]
   \caption{Adaptive Accelerated Algorithms with Clipping}
   \label{alg:thnntn4}
   \textbf{Input:} initial point $\mathbf{w}_1 = \tilde{\mathbf{w}}_1$, $\alpha_t \in (0,1]$, step sizes $\{ \eta_t \}_t$ and $\{ \beta_t \}_t$, dataset $S = \{z_1,...,z_n \}$, $G_0>0$, and $\tau > 0$.\\
       \vspace{-0.4cm}
\begin{algorithmic}[1]
    \FOR{$t=1,...,T$}
    \STATE  draw  $j_t$ from the uniform distribution over the set $\{ j: j\in [n] \}$ \\
          \STATE obtain $\bar{\mathbf{w}}_t = \alpha_t \mathbf{w}_t + (1-  \alpha_t) \tilde{\mathbf{w}}_t$ \\
   \STATE obtain $\nabla \bar{f}(\bar{\mathbf{w}}_t;z_{j_t})  = \frac{\nabla f(\bar{\mathbf{w}}_t;z_{j_t})}{\|\nabla f(\bar{\mathbf{w}}_t;z_{j_t})\|}\min \{ \tau, \|\nabla f(\bar{\mathbf{w}}_t;z_{j_t})\| \}$ \\
   \STATE update $\mathbf{w}_{t+1} = \mathbf{w}_{t} - \eta_t \nabla \bar{f}(\bar{\mathbf{w}}_t;z_{j_t})$\\
      \STATE update $\tilde{\mathbf{w}}_{t+1} = \bar{\mathbf{w}}_{t} - \beta_t \nabla \bar{f}(\bar{\mathbf{w}}_t;z_{j_t})$.
   \ENDFOR
\end{algorithmic}
\end{algorithm}
We then study a general form of adaptive accelerated algorithm, Algorithm \ref{alg:thnntn4}, which corresponds to a clipped version of Algorithm 2 in \cite{kavis2021high}. We introduce some adaptive algorithms covered by Algorithm \ref{alg:thnntn4}. Define $\lambda_t = \frac{1}{\sqrt{G_0^2 + \sum_{k=1}^t \|\nabla \bar{f}(\bar{\mathbf{w}}_t;z_{j_t})\|^2} }$. When $\eta_t =\beta_t= \lambda_t $, Algorithm \ref{alg:thnntn4} becomes the clipped AdaGrad. 
When $\eta_t = \lambda_t $ and $\beta_t = (1+\alpha_t) \eta_t$, where $\alpha_t = \frac{2}{t+1}$, Algorithm \ref{alg:thnntn4} becomes the clipped RSAG \cite{ghadimi2016accelerated}. Note that for Algorithm \ref{alg:thnntn4}, we are interested in the iterate $\bar{\mathbf{w}}_t$. Assumption \ref{brhbr} should be assumed on $\bar{\mathbf{w}}$, i.e., $\mathbb{E}_{j_t}[\|\nabla f(\bar{\mathbf{w}}_t;z_{j_t})\|^{\alpha}] \leq  G^{\alpha}$. We present the optimization guarantee below.
\begin{thm}\label{theorem7777}
Suppose the empirical risk $F_S$ satisfies Assumption \ref{ass1} and suppose Assumption \ref{brhbr} holds.  Assume that $F_S(\mathbf{w}) \leq M$ for all $\mathbf{w}$ for some $M$. Let $\bar{\mathbf{w}}_t$ be the iterate produced by Algorithm \ref{alg:thnntn4}.  Set $\tau = p T^{\frac{1}{3\alpha - 2}}$ for some positive constants $p$ such that $p\leq T^{\frac{2\alpha-2}{\alpha(3\alpha-2)}}$.
Then for any $\delta \in (0,1)$, with probability $1-\delta$, we have
\begin{align*}
\frac{1}{T}\sum_{t=1}^T \| \nabla F_S(\bar{\mathbf{w}}_{t}) \|^2
= \mathcal{O}\left(\frac{1}{T^\frac{2\alpha -2}{3\alpha - 2}}\log \frac{1}{\delta}\right).
\end{align*}
\end{thm}
\begin{remark}\rm{}
Algorithm \ref{alg:thnntn4} corresponds to a specific reformulation of Nesterov's acceleration \cite{ghadimi2016accelerated}.  
This reformulation was referred to as linear coupling in \cite{allen2014linear}, which is a combination of mirror descent, SGD, and averaging.
Theorem \ref{theorem7777} shows a similar $\mathcal{O}(\log \frac{1}{\delta}/T^\frac{2\alpha -2}{3\alpha - 2})$ rate to Theorem \ref{thoremss}. When $\alpha = 2$, it implies $\mathcal{O}(\log (\frac{1}{\delta})/T^\frac{1}{2})$. In the related work, \cite{kavis2021high} provide a convergence rate of the order $\mathcal{O}(\log (1/\delta)/\sqrt{T})$ for Algorithm \ref{alg:thnntn4} without clipping by assuming the smoothness, Lipschitz continuity of $F_S$, bounded variance, and $\|\nabla f(\mathbf{w}_t;z_{j_t})\| \leq G$ holding for all $\mathbf{w}_t$ and $z_{j_t}$ almost surely. By comparison, Theorem \ref{theorem7777} gives the guarantee for a heavy-tailed assumption allowing unbounded variance, and the overall conditions are weaker than \cite{kavis2021high}.
\end{remark}
\subsection{Summary of Results}
\begin{table*}[tb]
\caption{Summary of Results.}
\label{sample-table}
\vskip 0.15in
\centering
\begin{small}
\begin{sc}
\begin{tabular}{lccccr}
\toprule
Ref. & Algorithm & Assumption &  Measure &  Guarantee\\
\midrule
      \multirow{2}{*}{\cite{cutkosky2021high}}&\multirow{2}{*}{SGDM}  &  S, $\alpha$   &   $\frac{1}{T}\sum_{t=1}^T\| \nabla F_S(\mathbf{w}_{t}) \|$ &   $\mathcal{O} \Big(\frac{\log(T/\delta) }{T^{\frac{\alpha-1}{3\alpha-2}}}  \Big)$\\ 
&   &  S, $\alpha$, S-S   &  $\frac{1}{T}\sum_{t=1}^T\| \nabla F_S(\mathbf{w}_{t}) \|$  &    $\mathcal{O} \Big(\frac{\log(T/\delta) }{T^{\frac{2\alpha-2}{5\alpha-3}}}  \Big)$
  \\ \hline
       \multirow{7}{*}{Ours} &\multirow{2}{*}{SGD} &  S, $\alpha$  & $\frac{1}{T}\sum_{t=1}^T\| \nabla F_S(\mathbf{w}_{t}) \|^2$ & $\mathcal{O} \Big(\frac{1}{T^\frac{2\alpha -2}{3\alpha - 2}}\log \frac{1}{\delta}  \Big)$\\
             &  & S, $\alpha$ &$\frac{1}{T}\sum_{t=1}^T\| \nabla F(\mathbf{w}_{t}) \|^2$ & $\mathcal{O}\Big(\big(\frac{d}{n}\big)^{\frac{1}{2}}\log \frac{1}{\delta}\Big)$\\ 
 &\multirow{2}{*}{SGDM} & S, $\alpha$ & $\frac{1}{T}\sum_{t=1}^T\| \nabla F_S(\mathbf{w}_{t}) \|$ & $\mathcal{O} \Big(\frac{1}{T^\frac{\alpha -1}{3\alpha - 2}}\log \frac{T}{\delta} \Big)$\\ 
 &  &  S, $\alpha$ & $\frac{1}{T}\sum_{t=1}^T\| \nabla F(\mathbf{w}_{t}) \|$ & $\mathcal{O}\Big(\big(\frac{d}{n}\big)^{\frac{1}{4}}\log \frac{n}{d\delta}\Big)$\\
&\multirow{2}{*}{AdaGrad}                & S, $\alpha$ & $\frac{1}{T}\sum_{t=1}^T\| \nabla F_S(\mathbf{w}_{t}) \|^2$ & $\mathcal{O} \Big(\frac{1}{T^\frac{2\alpha -2}{3\alpha - 2}}\log \frac{1}{\delta}\Big)$\\
 &  & S, $\alpha$ & $\frac{1}{T}\sum_{t=1}^T\| \nabla F(\mathbf{w}_{t}) \|^2$ & $\mathcal{O} \Big( \Big(\frac{d}{n}\Big)^{\frac{2\alpha-2}{5\alpha-4}} \log \frac{1}{\delta} \log (1+(\frac{n}{d})^{\frac{3\alpha}{5\alpha-4}})\Big)$\\
 &Algorithm \ref{alg:thnntn4} & S, $\alpha$ & $\frac{1}{T}\sum_{t=1}^T\| \nabla F_S(\bar{\mathbf{w}}_{t}) \|^2$ & $\mathcal{O} \Big( \frac{1}{T^\frac{2\alpha -2}{3\alpha - 2}}\log \frac{1}{\delta} \Big)$\\ 
\bottomrule
\end{tabular}
\end{sc}
\end{small}
\vskip -0.1in
\end{table*}
We provide the results obtained in this paper and the high probability results of related work in the non-convex setting with gradient clipping in Table \ref{sample-table}.
Here, we provide some descriptions of Table \ref{sample-table}. S means the smoothness, S-S means second-order smoothness, and $\alpha$ means Assumption \ref{brhbr}. We say a function $g$ is $\rho$-second-order smoothness if for every $\mathbf{w}_1, \mathbf{w}_2 \in \mathcal{W}$ and $\mathbf{y} \in \mathbb{R}^d$, there holds
\begin{align*}
\| (\nabla^2 g(\mathbf{w}_1) -  \nabla^2 g(\mathbf{w}_2) ) \mathbf{y}\|^2 \leq \rho \| \mathbf{w}_1 - \mathbf{w}_2 \| \|  \mathbf{y}\|.
\end{align*}
One can derive the convergence bound and generalization bound for Algorithm \ref{alg:gdcil} with the second-order smoothness by incorporating our proof technique and the technique of \cite{cutkosky2021high}. We leave it to the interested readers. 
The difference between ours and \cite{cutkosky2021high} has been discussed in Remark \ref{remarks}.

Moreover, the comparison between our results and the results of related work (in-expectation analysis and high probability analysis) has been discussed in previous Remarks. We won't repeat it here and only provide an intuitive display of the related results here. One can see from Table \ref{sample-table} that we have provided a series of high probability convergence bounds and high probability generalization bounds for non-convex stochastic optimization with clipping that the related work does not involve.
\section{Conclusions}\label{concousion}
This paper provides a high probability analysis for non-convex stochastic optimization with clipping. We establish learning guarantees for clipped SGD and its variants of momentum and adaptive stepsizes under a heavy-tailed assumption of the stochastic gradients. Our analysis involves joint consideration of optimization and generalization performance, which systematically demonstrates the learning guarantees of non-convex stochastic optimization with gradient clipping from the two perspectives, and covers many popular stochastic optimization algorithms. We believe our theoretical findings can provide deep insights into the theoretical properties of stochastic optimization with clipping.

\section{Auxiliary Lemmas}\label{seclemma}
The following Lemma \ref{lemma5} and Lemma \ref{lemma51} provide concentration inequalities for martingales.
\begin{lem}[\cite{zhang2005data}]\label{lemma5}
Let $z_1,...,z_n$ be a sequence of randoms variables such that $z_k$ may depend the previous variables $z_1,...,z_{k-1}$ for all $k = 1,...,n$. Consider a sequence of functionals $\xi_k(z_1,...,z_k)$, $k=1,...,n$. Let $\sigma_n^2 = \sum_{k=1}^n \mathbb{E}_{z_k} [ (\xi_k - \mathbb{E}_{z_k}[\xi_k])^2]$ be the conditional variance.
Assume $|\xi_k - \mathbb{E}_{z_k}[\xi_k] | \leq b$ for each $k$. Let $\rho \in (0,1]$ and $\delta \in (0,1)$. With probability at least $1-\delta$ we have 
\begin{align*}
\sum_{k=1}^n \xi_k - \sum_{k=1}^n \mathbb{E}_{z_k}[\xi_k] \leq \frac{\rho \sigma_n^2}{b} + \frac{b \log \frac{1}{\delta}}{\rho}.
\end{align*}
\end{lem}
\begin{lem}[\cite{tarres2014online}]\label{lemma51}
Let $\{\xi_k \}_{k \in \mathbb{N}}$ be a martingale difference sequence in $\mathbb{R}^d$. Suppose that almost surely $\| \xi_k  \| \leq D$ and $ \sum_{k=1}^t \mathbb{E} [ \| \xi_k \|^2 | \xi_1,...,\xi_{k-1}]\leq \sigma_t^2$. Then, for any $0< \delta < 1$, the following inequality holds with probability at least $1-\delta$
\begin{align*}
\max_{1\leq j \leq t} \left \| \sum_{k =1}^j \xi_k \right\| \leq 2\left( \frac{D}{3} + \sigma_t  \right)\log \frac{2}{\delta}.
\end{align*}
\end{lem}
The following Lemma \ref{lemmab3} states the uniform convergence of the gradient, which will be used to derive the generalization bound of this paper.
\begin{lem}[\cite{lei2021learning}]\label{lemmab3}
Let $\delta \in (0,1)$, $R > 0$, and $S= \{z_1,...,z_n  \}$ be a set of i.i.d. samples. Suppose the function $f$ satisfies Assumption \ref{ass1}. Then with probability at least $1-\delta$ we have
\begin{align*}
&\sup_{\mathbf{w} \in B_R} \| \nabla F(\mathbf{w}) -  \nabla F_S(\mathbf{w})\|\\&\leq \frac{(L R + b)}{\sqrt{n}}\Big(2+ 2\sqrt{48e\sqrt{2}(\log 2 + d \log(3e))} + \sqrt{2 \log (\frac{1}{\delta})}\Big),
\end{align*}
where $e$ is the base of the natural logarithm.
\end{lem}

The following Lemma \ref{lemma8} is from online learning and is often used in the study of adaptive algorithms \cite{kavis2021high,ward2019adagrad,li2019convergence,li2020high}.
\begin{lem}\label{lemma8}
Let $a_1,...,a_n$ be a sequence of non-negative real numbers. Then, it holds that 
\begin{align*}
\sqrt{\sum_{i=1}^na_i} \leq \sum_{i=1}^n \frac{a_i}{\sqrt{\sum_{k=1}^ia_k}}\leq 2\sqrt{\sum_{i=1}^na_i},
\end{align*}
and
\begin{align*}
\sum_{i=1}^n \frac{a_i}{\sum_{k=1}^ia_k}\leq 1+\log\Big(1+\sum_{i=1}^na_i\Big).
\end{align*}
\end{lem}

\bibliography{ecai}
\appendix
\section{Proofs}\label{proof6}

\subsection{Proof of Theorem 5}\label{appendixbi.}
\begin{proof}
It is easy to verify that $\|\nabla \bar{f}(\mathbf{w}_t;z_{j_t}) \| \leq \tau$. With the descent lemma of smoothness and $\eta_t = \eta$, we have
\begin{align*}
&F_S(\mathbf{w}_{t+1}) - F_S(\mathbf{w}_{t}) \\
&\leq \langle \mathbf{w}_{t+1} - \mathbf{w}_{t}, \nabla F_S(\mathbf{w}_{t}) \rangle + \frac{L}{2}  \| \mathbf{w}_{t+1} -\mathbf{w}_{t}  \|^2\\
& \leq -\eta \langle  \nabla \bar{f}(\mathbf{w}_t;z_{j_t}), \nabla F_S(\mathbf{w}_{t}) \rangle + \frac{L}{2}  \eta^2 \|\nabla \bar{f}(\mathbf{w}_t;z_{j_t}) \|^2 \\
& = -\eta    \langle \nabla \bar{f}(\mathbf{w}_t;z_{j_t}) - \mathbb{E}_{j_t}\nabla \bar{f}(\mathbf{w}_t;z_{j_t}) \\
&+ \mathbb{E}_{j_t}\nabla \bar{f}(\mathbf{w}_t;z_{j_t}) -  \nabla F_S(\mathbf{w}_{t})  , \nabla F_S(\mathbf{w}_{t}) \rangle - \eta \| \nabla F_S(\mathbf{w}_{t}) \|^2 \\&+  \frac{L}{2}  \eta^2 \|\nabla \bar{f}(\mathbf{w}_t;z_{j_t}) - \mathbb{E}_{j_t}\nabla \bar{f}(\mathbf{w}_t;z_{j_t}) \\
&+ \mathbb{E}_{j_t}\nabla \bar{f}(\mathbf{w}_t;z_{j_t}) -  \nabla F_S(\mathbf{w}_{t}) + \nabla F_S(\mathbf{w}_{t}) \|^2  \\
& \leq -\eta    \langle \nabla \bar{f}(\mathbf{w}_t;z_{j_t}) - \mathbb{E}_{j_t}\nabla \bar{f}(\mathbf{w}_t;z_{j_t}), \nabla F_S(\mathbf{w}_{t}) \rangle   \\
&-\eta \langle\mathbb{E}_{j_t}\nabla \bar{f}(\mathbf{w}_t;z_{j_t}) -  \nabla F_S(\mathbf{w}_{t})  , \nabla F_S(\mathbf{w}_{t}) \rangle \\&- \eta \| \nabla F_S(\mathbf{w}_{t}) \|^2 +  \frac{3}{2} L \eta^2 \|\nabla \bar{f}(\mathbf{w}_t;z_{j_t}) - \mathbb{E}_{j_t}\nabla \bar{f}(\mathbf{w}_t;z_{j_t})\|^2\\
&-  \frac{3}{2} L \eta^2 \mathbb{E}_{j_t}\|\nabla \bar{f}(\mathbf{w}_t;z_{j_t}) - \mathbb{E}_{j_t}\nabla \bar{f}(\mathbf{w}_t;z_{j_t})\|^2 \\&+  \frac{3}{2} L \eta^2 \mathbb{E}_{j_t}\|\nabla \bar{f}(\mathbf{w}_t;z_{j_t}) - \mathbb{E}_{j_t}\nabla \bar{f}(\mathbf{w}_t;z_{j_t})\|^2 \\
&+  \frac{3}{2} L \eta^2\|\mathbb{E}_{j_t}\nabla \bar{f}(\mathbf{w}_t;z_{j_t}) -  \nabla F_S(\mathbf{w}_{t})\|^2 +  \frac{3}{2} L \eta^2\|\nabla F_S(\mathbf{w}_{t}) \|^2.
\end{align*}
By a summation from $t=1$ to $t=T$ and according to $\eta \leq 1/(12L)$, we get 
\begin{align}\label{thefirst}\nonumber
&\frac{7}{8}\sum_{t=1}^T \eta \| \nabla F_S(\mathbf{w}_{t}) \|^2 \leq F_S(\mathbf{w}_{1}) - F_S(\mathbf{w}_S^{\ast}) \\\nonumber
& +\sum_{t=1}^T \frac{3}{2} L \eta^2 (\|\nabla \bar{f}(\mathbf{w}_t;z_{j_t}) - \mathbb{E}_{j_t}\nabla \bar{f}(\mathbf{w}_t;z_{j_t})\|^2\\\nonumber
&-\mathbb{E}_{j_t}\|\nabla \bar{f}(\mathbf{w}_t;z_{j_t}) - \mathbb{E}_{j_t}\nabla \bar{f}(\mathbf{w}_t;z_{j_t})\|^2)\\\nonumber& + \sum_{t=1}^T \frac{3}{2} L \eta^2\|\mathbb{E}_{j_t}\nabla \bar{f}(\mathbf{w}_t;z_{j_t}) -  \nabla F_S(\mathbf{w}_{t})\|^2 \\\nonumber
&+ \sum_{t=1}^T\frac{3}{2} L \eta^2 \mathbb{E}_{j_t}\|\nabla \bar{f}(\mathbf{w}_t;z_{j_t}) - \mathbb{E}_{j_t}\nabla \bar{f}(\mathbf{w}_t;z_{j_t})\|^2 \\\nonumber&- \sum_{t=1}^T  \eta    \langle \nabla \bar{f}(\mathbf{w}_t;z_{j_t}) - \mathbb{E}_{j_t}\nabla \bar{f}(\mathbf{w}_t;z_{j_t}), \nabla F_S(\mathbf{w}_{t}) \rangle \\
&-  \sum_{t=1}^T \eta\langle\mathbb{E}_{j_t}\nabla \bar{f}(\mathbf{w}_t;z_{j_t}) -  \nabla F_S(\mathbf{w}_{t})  , \nabla F_S(\mathbf{w}_{t}) \rangle .
\end{align}

Firstly, $F_S(\mathbf{w}_{1}) - F_S(\mathbf{w}_S^{\ast})$ can be seen as a constant.
Since $ \mathbb{E}_{j_t}[ -\eta  \langle \nabla \bar{f}(\mathbf{w}_t;z_{j_t}) - \mathbb{E}_{j_t}\nabla \bar{f}(\mathbf{w}_t;z_{j_t}), \nabla F_S(\mathbf{w}_{t}) \rangle ]= 0$, the sequence $\{-\eta    \langle \nabla \bar{f}(\mathbf{w}_t;z_{j_t}) - \mathbb{E}_{j_t}\nabla \bar{f}(\mathbf{w}_t;z_{j_t}), \nabla F_S(\mathbf{w}_{t}) \rangle, t \in \mathbb{N} \}$ is a martingale difference sequence. For brevity, denoted by $\xi_t = -\eta    \langle \nabla \bar{f}(\mathbf{w}_t;z_{j_t}) - \mathbb{E}_{j_t}\nabla \bar{f}(\mathbf{w}_t;z_{j_t}), \nabla F_S(\mathbf{w}_{t}) \rangle$. 

Besides, there holds
\begin{align}\label{ineq00ji}\nonumber
\| \nabla F_S(\mathbf{w}_{t}) \|^{\alpha}&= \|\mathbb{E}_{j_t}[\nabla f(\mathbf{w}_t;z_{j_t})]\|^{\alpha}\\
& \leq \mathbb{E}_{j_t}[\|\nabla f(\mathbf{w}_t;z_{j_t})\|^{\alpha}] \leq  G^{\alpha},
\end{align}
where the first inequality follows from the Jensen's inequality and the second inequality follows from Assumption 2.
We thus have $\| \nabla F_S(\mathbf{w}_{t}) \| \leq G$. 

Further, we can derive that
\begin{align*}
|\xi_t| &\leq \eta  (\|  \nabla \bar{f}(\mathbf{w}_t;z_{j_t}) \| + \| \mathbb{E}_{j_t}\nabla \bar{f}(\mathbf{w}_t;z_{j_t})\| )\| \nabla F_S(\mathbf{w}_{t}) \| \\
&\leq 2\eta \tau G,
\end{align*}
and that
\begin{align}\label{eq789}\nonumber
 &\mathbb{E}_{j_t} [\| \nabla \bar{f}(\mathbf{w}_t;z_{j_t}) - \mathbb{E}_{j_t}\nabla \bar{f}(\mathbf{w}_t;z_{j_t})\|^2 ] \\\nonumber&\leq  \mathbb{E}_{j_t} [\| \nabla \bar{f}(\mathbf{w}_t;z_{j_t}) \|^2 ]\\\nonumber& = \mathbb{E}_{j_t} [\| \nabla \bar{f}(\mathbf{w}_t;z_{j_t}) \|^\alpha \| \nabla \bar{f}(\mathbf{w}_t;z_{j_t}) \|^{2-\alpha} ]\\& \leq \mathbb{E}_{j_t} [\| \nabla f(\mathbf{w}_t;z_{j_t}) \|^\alpha \tau^{2-\alpha}] \leq G^\alpha \tau^{2-\alpha},
\end{align}
where the first inequality holds according the property of variance, the second inequality holds due to $\| \nabla \bar{f}(\mathbf{w}_t;z_{j_t}) \| \leq \| \nabla f(\mathbf{w}_t;z_{j_t}) \|$, and the third inequality follows from Assumption 2. Further,
according this property, we have $\mathbb{E}_{j_t} [(\xi_t - \mathbb{E}_{j_t} \xi_t)^2] \leq \mathbb{E}_{j_t} [\xi_t^2]$ and thus have 
\begin{align}\label{eqzuihouyigeshizi}\nonumber
&\sum_{t=1}^T  \mathbb{E}_{j_t} [(\xi_t - \mathbb{E}_{j_t} \xi_t)^2]\\\nonumber&\leq \sum_{t=1}^T \eta^2  \mathbb{E}_{j_t} [\| \nabla \bar{f}(\mathbf{w}_t;z_{j_t}) - \mathbb{E}_{j_t}\nabla \bar{f}(\mathbf{w}_t;z_{j_t})\|^2 \| \nabla F_S(\mathbf{w}_{t}) \|^2] \\&\leq G^\alpha \tau^{2-\alpha} \sum_{t=1}^T \eta^2   \|\nabla F_S(\mathbf{w}_{t}) \|^2.
\end{align}
According to Lemma 19, with probability $1- \delta$, we have
\begin{align*}
\sum_{t=1}^T \xi_t \leq \frac{\rho G^\alpha \tau^{2-\alpha} \eta \sum_{t=1}^T \eta   \|\nabla F_S(\mathbf{w}_{t}) \|^2}{2\eta \tau G} + \frac{2\eta \tau G \log(1/\delta)}{\rho}.
\end{align*}
Taking $\rho = \min\left\{\frac{G^{1-\alpha}}{2\tau^{1-\alpha}},1\right\} $, we get with probability $1- \delta$
\begin{align}\label{inequbuzhidao}\nonumber
&\sum_{t=1}^T \xi_t \leq \frac{\sum_{t=1}^T \eta   \|\nabla F_S(\mathbf{w}_{t}) \|^2}{4} \\&+ \max\{4\eta \tau^{2-\alpha} G^{\alpha}, 2\eta \tau G\} \log(\frac{1}{\delta}).
\end{align}

Next, we bound the term $-  \sum_{t=1}^T \eta\langle\mathbb{E}_{j_t}\nabla \bar{f}(\mathbf{w}_t;z_{j_t}) -  \nabla F_S(\mathbf{w}_{t})  , \nabla F_S(\mathbf{w}_{t}) \rangle$. We have
\begin{align*}
&-  \sum_{t=1}^T \eta\langle\mathbb{E}_{j_t}\nabla \bar{f}(\mathbf{w}_t;z_{j_t}) -  \nabla F_S(\mathbf{w}_{t})  , \nabla F_S(\mathbf{w}_{t}) \rangle \\
&\leq \frac{1}{2}\sum_{t=1}^T \eta \| \mathbb{E}_{j_t}\nabla \bar{f}(\mathbf{w}_t;z_{j_t}) -  \nabla F_S(\mathbf{w}_{t})  \|^2 + \frac{1}{2}\sum_{t=1}^T \eta  \| \nabla F_S(\mathbf{w}_{t}) \|^2.
\end{align*}
Denoted by $x_t = \mathbb{I}_{\{ \| \nabla f(\mathbf{w}_t;z_{j_t}) \| > \tau\}}$.
For the term $\| \mathbb{E}_{j_t}\nabla \bar{f}(\mathbf{w}_t;z_{j_t}) -  \nabla F_S(\mathbf{w}_{t})  \| $, we have 
\begin{align}\label{eq111}\nonumber
&\| \mathbb{E}_{j_t}\nabla \bar{f}(\mathbf{w}_t;z_{j_t}) -  \nabla F_S(\mathbf{w}_{t})  \| \\\nonumber&= \| \mathbb{E}_{j_t}\Big [(\nabla \bar{f}(\mathbf{w}_t;z_{j_t}) -\nabla f(\mathbf{w}_t;z_{j_t}) )\Big ]\|\\\nonumber
&= \|  \mathbb{E}_{j_t}  \Big [\nabla f(\mathbf{w}_t;z_{j_t})(  \frac{\tau}{\| \nabla f(\mathbf{w}_t;z_{j_t}) \|} -1 ) x_t \Big ] \| \\\nonumber
&\leq \mathbb{E}_{j_t}\Big [ \|   \nabla f(\mathbf{w}_t;z_{j_t})  x_t \| \Big ]\leq \mathbb{E}_{j_t}  \Big [\| \nabla f(\mathbf{w}_t;z_{j_t})\|^\alpha \tau^{1-\alpha}\Big ] \\
&\leq G^\alpha \tau^{1-\alpha},
\end{align}
where the first inequality holds due to Jensen's inequality and that $ \frac{\tau}{\| \nabla f(\mathbf{w}_t;z_{j_t}) \|} \in (0,1)$, the second inequality holds due to $ \| \nabla f(\mathbf{w}_t;z_{j_t}) \| > \tau$, and the last inequality holds due to Assumption 2.
Thus, we have the following inequality 
\begin{align}\label{eq999}\nonumber
&-  \sum_{t=1}^T \eta\langle\mathbb{E}_{j_t}\nabla \bar{f}(\mathbf{w}_t;z_{j_t}) -  \nabla F_S(\mathbf{w}_{t})  , \nabla F_S(\mathbf{w}_{t}) \rangle \\
\leq &\frac{1}{2}G^{2\alpha} \tau^{2-2\alpha} T \eta  + \frac{1}{2}\sum_{t=1}^T \eta  \| \nabla F_S(\mathbf{w}_{t}) \|^2.
\end{align}

Plugging (\ref{inequbuzhidao}) and (\ref{eq999}) into (\ref{thefirst}) and combining the fact $\| \mathbb{E}_{j_t}\nabla \bar{f}(\mathbf{w}_t;z_{j_t}) -  \nabla F_S(\mathbf{w}_{t})  \| \leq G^\alpha \tau^{1-\alpha}$, we have the following inequality with probability $1 - \delta$
\begin{align*}
&\frac{1}{8}\sum_{t=1}^T \eta\| \nabla F_S(\mathbf{w}_{t}) \|^2 
\leq F_S(\mathbf{w}_{1}) - F_S(\mathbf{w}_S) \\&+\sum_{t=1}^T \frac{3}{2} L \eta^2 \|\nabla \bar{f}(\mathbf{w}_t;z_{j_t}) - \mathbb{E}_{j_t}\nabla \bar{f}(\mathbf{w}_t;z_{j_t})\|^2\\&-\mathbb{E}_{j_t}\|\nabla \bar{f}(\mathbf{w}_t;z_{j_t}) - \mathbb{E}_{j_t}\nabla \bar{f}(\mathbf{w}_t;z_{j_t})\|^2\\& + \sum_{t=1}^T\frac{3}{2} L \eta^2 \mathbb{E}_{j_t}\|\nabla \bar{f}(\mathbf{w}_t;z_{j_t}) - \mathbb{E}_{j_t}\nabla \bar{f}(\mathbf{w}_t;z_{j_t})\|^2\\& + \frac{3}{2} L \eta^2  G^{2\alpha} \tau^{2-2\alpha}T + \max\{4\eta \tau^{2-\alpha} G^{\alpha}, 2\eta \tau G\} \log(1/\delta)\\& + \frac{1}{2}G^{2\alpha} \tau^{2-2\alpha}T \eta.
\end{align*}
Moreover, there holds that
\begin{align*}
&\sum_{t=1}^T \mathbb{E}_{j_t} \|\nabla \bar{f}(\mathbf{w}_t;z_{j_t}) - \mathbb{E}_{j_t}\nabla \bar{f}(\mathbf{w}_t;z_{j_t})\|^2\\&\leq \sum_{t=1}^T \mathbb{E}_{j_t} \|\nabla \bar{f}(\mathbf{w}_t;z_{j_t})\|^2\\&\leq \sum_{t=1}^T \mathbb{E}_{j_t} \|\nabla \bar{f}(\mathbf{w}_t;z_{j_t})\|^\alpha  \|\nabla \bar{f}(\mathbf{w}_t;z_{j_t})\|^{2-\alpha}\leq  TG^\alpha\tau^{2-\alpha}.
\end{align*}
To proceed, we now bound the term $\sum_{t=1}^T \|\nabla \bar{f}(\mathbf{w}_t;z_{j_t}) - \mathbb{E}_{j_t}\nabla \bar{f}(\mathbf{w}_t;z_{j_t})\|^2-\mathbb{E}_{j_t}\|\nabla \bar{f}(\mathbf{w}_t;z_{j_t}) - \mathbb{E}_{j_t}\nabla \bar{f}(\mathbf{w}_t;z_{j_t})\|^2$. Since $ \mathbb{E}_{j_t}[ \|\nabla \bar{f}(\mathbf{w}_t;z_{j_t}) - \mathbb{E}_{j_t}\nabla \bar{f}(\mathbf{w}_t;z_{j_t})\|^2-\mathbb{E}_{j_t}\|\nabla \bar{f}(\mathbf{w}_t;z_{j_t}) - \mathbb{E}_{j_t}\nabla \bar{f}(\mathbf{w}_t;z_{j_t})\|^2]= 0$, the sequence $\{\|\nabla \bar{f}(\mathbf{w}_t;z_{j_t}) - \mathbb{E}_{j_t}\nabla \bar{f}(\mathbf{w}_t;z_{j_t})\|^2-\mathbb{E}_{j_t}\|\nabla \bar{f}(\mathbf{w}_t;z_{j_t}) - \mathbb{E}_{j_t}\nabla \bar{f}(\mathbf{w}_t;z_{j_t})\|^2, t \in \mathbb{N} \}$ is a martingale difference sequence. 

For any $t \in \mathbb{N}$, we have the following inequality  almost surely
\begin{align*}
&\|\nabla \bar{f}(\mathbf{w}_t;z_{j_t}) - \mathbb{E}_{j_t}\nabla \bar{f}(\mathbf{w}_t;z_{j_t})\|^2\\&-\mathbb{E}_{j_t}\|\nabla \bar{f}(\mathbf{w}_t;z_{j_t}) - \mathbb{E}_{j_t}\nabla \bar{f}(\mathbf{w}_t;z_{j_t})\|^2\leq 4\tau^2.
\end{align*}
We now consider the conditional variance.
We have 
\begin{align*}
&\sum_{t=1}^T\mathbb{E}_{j_t}\Big[\|\nabla \bar{f}(\mathbf{w}_t;z_{j_t}) - \mathbb{E}_{j_t}\nabla \bar{f}(\mathbf{w}_t;z_{j_t})\|^2\\
&-\mathbb{E}_{j_t}\|\nabla \bar{f}(\mathbf{w}_t;z_{j_t}) - \mathbb{E}_{j_t}\nabla \bar{f}(\mathbf{w}_t;z_{j_t})\|^2 \Big]^2 \\
&\leq \sum_{t=1}^T \mathbb{E}_{j_t}[\|\nabla \bar{f}(\mathbf{w}_t;z_{j_t}) - \mathbb{E}_{j_t}\nabla \bar{f}(\mathbf{w}_t;z_{j_t})\|^2]^2\\
&\leq 4\tau^2\sum_{t=1}^T \mathbb{E}_{j_t}[\|\nabla \bar{f}(\mathbf{w}_t;z_{j_t}) - \mathbb{E}_{j_t}\nabla \bar{f}(\mathbf{w}_t;z_{j_t})\|^2]\\
&\leq 4\tau^2\sum_{t=1}^T \mathbb{E}_{j_t} \|\nabla \bar{f}(\mathbf{w}_t;z_{j_t})\|^2\\&\leq 4\tau^2\sum_{t=1}^T \mathbb{E}_{j_t} \|\nabla \bar{f}(\mathbf{w}_t;z_{j_t})\|^\alpha  \|\nabla \bar{f}(\mathbf{w}_t;z_{j_t})\|^{2-\alpha}\leq  4TG^\alpha\tau^{4-\alpha},
\end{align*}
where the first and third inequalities hold due to the property of variance.

According to Lemma 19, we have the following inequality  with probability at least $1-\delta$
\begin{align*}
&\sum_{t=1}^T \Big[\|\nabla \bar{f}(\mathbf{w}_t;z_{j_t}) - \mathbb{E}_{j_t}\nabla \bar{f}(\mathbf{w}_t;z_{j_t})\|^2\\
&-\mathbb{E}_{j_t}\|\nabla \bar{f}(\mathbf{w}_t;z_{j_t}) - \mathbb{E}_{j_t}\nabla \bar{f}(\mathbf{w}_t;z_{j_t})\|^2\Big] \\
&\leq \frac{\rho 4TG^\alpha\tau^{4-\alpha} }{4\tau^2} +\frac{4\tau^2\log(1/\delta)}{\rho}.
\end{align*}
Taking $\rho = \frac{\tau^{\frac{\alpha}{2}}}{\sqrt{T}}$, we need $\frac{\tau^{\frac{\alpha}{2}}}{\sqrt{T}} \leq 1$, that is we need $\tau \leq  T^{\frac{1}{\alpha}}$.
We get
\begin{align*}
&\sum_{t=1}^T \Big[\|\nabla \bar{f}(\mathbf{w}_t;z_{j_t}) - \mathbb{E}_{j_t}\nabla \bar{f}(\mathbf{w}_t;z_{j_t})\|^2\\
&-\mathbb{E}_{j_t}\|\nabla \bar{f}(\mathbf{w}_t;z_{j_t}) - \mathbb{E}_{j_t}\nabla \bar{f}(\mathbf{w}_t;z_{j_t})\|^2\Big] \\
&\leq \sqrt{T}\tau^{2-\frac{\alpha}{2}} G^\alpha +4 \sqrt{T}\tau^{2-\frac{\alpha}{2}}\log(1/\delta).
\end{align*}
Then we get the following inequality  with probability at least $1-2\delta$
\begin{align*}
&\frac{1}{8}\sum_{t=1}^T \eta\| \nabla F_S(\mathbf{w}_{t}) \|^2 
\leq F_S(\mathbf{w}_{1}) - F_S(\mathbf{w}_S) \\&+  \frac{3}{2}L\eta^2\left(\sqrt{T}\tau^{2-\frac{\alpha}{2}} G^\alpha +4 \sqrt{T}\tau^{2-\frac{\alpha}{2}}\log(1/\delta) \right) \\& + \frac{3}{2} L \eta^2 TG^\alpha\tau^{2-\alpha} + \frac{3}{2} L \eta^2  G^{2\alpha} \tau^{2-2\alpha}T \\&+ \max\{4\eta \tau^{2-\alpha} G^{\alpha}, 2\eta \tau G\} \log(1/\delta) + \frac{1}{2}G^{2\alpha} \tau^{2-2\alpha}T \eta,
\end{align*}
which implies that
\begin{align}\label{thelast}\nonumber
&\frac{1}{T}\sum_{t=1}^T \| \nabla F_S(\mathbf{w}_{t}) \|^2 
\leq \frac{8(F_S(\mathbf{w}_{1}) - F_S(\mathbf{w}_S^{\ast}))}{T\eta}  + 4G^{2\alpha} \tau^{2-2\alpha} \\\nonumber&+12L \eta T^{-1/2}(\tau^{2-\frac{\alpha}{2}} G^\alpha  +4 \tau^{2-\frac{\alpha}{2}}\log(1/\delta))+ 12 L \eta G^\alpha\tau^{2-\alpha}\\& +  12L\eta G^{2\alpha} \tau^{2-2\alpha} +\frac{8}{T} \max\{4 \tau^{2-\alpha} G^{\alpha}, 2 \tau G\} \log(1/\delta) .
\end{align}

Thus, we select $\eta= p(\frac{1}{T})^{\frac{\alpha}{3\alpha-2}}$ and $\tau = q T^{\frac{1}{3\alpha - 2}}$ for some positive constant $p,q >0$ such that $q\leq T^{\frac{2\alpha-2}{\alpha(3\alpha-2)}}$, where the constrain of $q$ holds due to the requirement $\tau \leq  T^{\frac{1}{\alpha}}$.

Till here, we finally obtain the following inequality  with probability $1-2\delta$
\begin{align*}
\frac{1}{T}\sum_{t=1}^T \| \nabla F_S(\mathbf{w}_{t}) \|^2 = \mathcal{O}\left(\frac{\log 1/\delta}{T^\frac{2\alpha -2}{3\alpha - 2}}\right).
\end{align*}
Note that the dependence on confidence parameter $1/\delta$ in the above inequality is logarithmic. One can replace $\delta$ to $\delta/2$. Hence,
the above inequality implies that with probability $1-\delta$
\begin{align*}
&\frac{1}{T}\sum_{t=1}^T \| \nabla F_S(\mathbf{w}_{t}) \|^2 = \mathcal{O}\left(\frac{\log 1/\delta}{T^\frac{2\alpha -2}{3\alpha - 2}}\right).
\end{align*}
The proof is complete.
\end{proof}
\subsection{Proof of Theorem 7}\label{appendixb2}
\begin{proof}
Since $\mathbf{w}_{t+1} -\mathbf{w}_{t} = \eta \nabla \bar{f}(\mathbf{w}_t;z_{j_t})$, using $\mathbf{w}_{1} = \mathbf{0}$, we have
\begin{align*}
&\| \mathbf{w}_{t+1} \| =  \eta\| \sum_{k=1}^t  \nabla \bar{f}(\mathbf{w}_k;z_{j_k}) \|\\
& =  \eta\| \sum_{k=1}^t  \nabla \bar{f}(\mathbf{w}_k;z_{j_k}) - \mathbb{E}_{j_k}\nabla \bar{f}(\mathbf{w}_k;z_{j_k}) \\
&+ \mathbb{E}_{j_k}\nabla \bar{f}(\mathbf{w}_k;z_{j_k}) -  \nabla F_S(\mathbf{w}_{k}) + \nabla F_S(\mathbf{w}_{k}) \|\\
&\leq  \eta\| \sum_{k=1}^t  \nabla \bar{f}(\mathbf{w}_k;z_{j_k}) - \mathbb{E}_{j_k}\nabla \bar{f}(\mathbf{w}_k;z_{j_k})\| \\
&+ \eta\| \sum_{k=1}^t\mathbb{E}_{j_k}\nabla \bar{f}(\mathbf{w}_k;z_{j_k}) -  \nabla F_S(\mathbf{w}_{k}) \|+  \eta\|\sum_{k=1}^t\nabla F_S(\mathbf{w}_{k}) \|.
\end{align*}
It is clear that the sequence $\{\nabla \bar{f}(\mathbf{w}_k;z_{j_k}) - \mathbb{E}_{j_k}\nabla \bar{f}(\mathbf{w}_k;z_{j_k}), t \in \mathbb{N} \}$ is a martingale difference sequence.
For any $t \in \mathbb{N}$, we have the following inequality  almost surely
\begin{align*}
\|\nabla \bar{f}(\mathbf{w}_k;z_{j_k}) - \mathbb{E}_{j_k}\nabla \bar{f}(\mathbf{w}_k;z_{j_k})\|\leq 2\tau.
\end{align*}
And we have
\begin{align}\label{eqxuyao}\nonumber
&\sum_{k=1}^t \mathbb{E}_{j_k} \|\nabla \bar{f}(\mathbf{w}_k;z_{j_k}) - \mathbb{E}_{j_k}\nabla \bar{f}(\mathbf{w}_k;z_{j_k})\|^2\\\nonumber&\leq \sum_{k=1}^t \mathbb{E}_{j_k} \|\nabla \bar{f}(\mathbf{w}_k;z_{j_k})\|^2\\\nonumber&\leq \sum_{k=1}^t \mathbb{E}_{j_k} \|\nabla \bar{f}(\mathbf{w}_k;z_{j_k})\|^\alpha  \|\nabla \bar{f}(\mathbf{w}_k;z_{j_k})\|^{2-\alpha}\\&\leq  tG^\alpha\tau^{2-\alpha}.
\end{align}
According to Lemma 20, we have the following inequality  with probability at least $1-\delta$
\begin{align*}
&\| \sum_{k=1}^t  \nabla \bar{f}(\mathbf{w}_k;z_{j_k}) - \mathbb{E}_{j_k}\nabla \bar{f}(\mathbf{w}_k;z_{j_k})\|\\& \leq 2\left(\frac{2\tau}{3} + t^{1/2}G^{\alpha/2}\tau^{\frac{2-\alpha}{2}}\right)\log(2/\delta).
\end{align*}
Furthermore, according to (\ref{eq111}), we have
\begin{align}\label{eqyouxuyao}\nonumber
&\| \sum_{k=1}^t\mathbb{E}_{j_k}\nabla \bar{f}(\mathbf{w}_k;z_{j_k}) -  \nabla F_S(\mathbf{w}_{k}) \|\\&\leq \sum_{k=1}^t\| \mathbb{E}_{j_k}\nabla \bar{f}(\mathbf{w}_k;z_{j_k}) -  \nabla F_S(\mathbf{w}_{k}) \|\leq G^\alpha \tau^{1-\alpha}t.
\end{align}
Besides, we have the following inequality with probability at least $1 - \delta$
\begin{align*}
&\|\sum_{k=1}^t\nabla F_S(\mathbf{w}_{k}) \|^2 \leq \left(\sum_{k=1}^t\|\nabla F_S(\mathbf{w}_{k}) \|\right)^2\\
&\leq \left(\sum_{k=1}^t\right) \left(\sum_{k=1}^t\|\nabla F_S(\mathbf{w}_{k}) \|^2\right)\\
&\leq t^2  \mathcal{O}\left(\frac{\log 1/\delta}{t^\frac{2\alpha -2}{3\alpha - 2}}\right)= \mathcal{O}\left(t^{\frac{4\alpha-2}{3\alpha-2}}\log 1/\delta\right),
\end{align*}
where the second inequality follows from the Schwarz's inequality and the third inequality follows from the result of Theorem 5.
The above inequality implies that with probability at least $1 - \delta$
\begin{align*}
\|\sum_{k=1}^t\nabla F_S(\mathbf{w}_{k}) \|= \mathcal{O}\left(t^{\frac{2\alpha-1}{3\alpha-2}}\log^{\frac{1}{2}} 1/\delta\right).
\end{align*}
Thus, combining these bounds, we have the following inequality with probability at least $1 - 2\delta$
\begin{align*}
&\| \mathbf{w}_{t+1} \| \leq  \eta\| \sum_{k=1}^t  \nabla \bar{f}(\mathbf{w}_k;z_{j_k}) - \mathbb{E}_{j_k}\nabla \bar{f}(\mathbf{w}_k;z_{j_k})\| \\&+ \eta\| \sum_{k=1}^t\mathbb{E}_{j_k}\nabla \bar{f}(\mathbf{w}_k;z_{j_k}) -  \nabla F_S(\mathbf{w}_{k}) \|+ \eta \|\sum_{k=1}^t\nabla F_S(\mathbf{w}_{k}) \|\\
&\leq 2 \eta\left(\frac{2\tau}{3} + t^{1/2}G^{\alpha/2}\tau^{\frac{2-\alpha}{2}} \right)\log(2/\delta) +  \eta G^\alpha \tau^{1-\alpha}t \\&+  \eta\mathcal{O}(t^{\frac{2\alpha-1}{3\alpha-2}}\log^{\frac{1}{2}} 1/\delta).
\end{align*}
Since $\eta= p\frac{1}{T^{\frac{\alpha}{3\alpha-2}}}$ and $\tau = q T^{\frac{1}{3\alpha - 2}}$, we have the following inequality with probability at least $1 - 2\delta$ uniformly for all $t = 1, ..., T$
\begin{align}\label{unilform}
\| \mathbf{w}_{t+1} \|& \leq  \mathcal{O}\left(T^{\frac{\alpha-1}{3\alpha-2}}\log^{\frac{1}{2}} 1/\delta\right).
\end{align}
For brevity, denoted by $\gamma = (2+ 2\sqrt{48e\sqrt{2}(\log 2 + d \log(3e))} + \sqrt{2 \log (\frac{1}{\delta})})$. Plugging the bound of $\| \mathbf{w}_{t+1} \|$ into Lemma 21, we have the following inequality with probability at least $1 - 3\delta$
uniformly for all $t = 1, ...T$
\begin{align}\label{eiglhelg}\nonumber
&\| \nabla F(\mathbf{w}_{t+1}) -  \nabla F_S(\mathbf{w}_{t+1})\|  \leq \frac{(L R_{t+1} + b)}{\sqrt{n}}\gamma\\
= &\frac{(L \|\mathbf{w}_{t+1}\| + b)}{\sqrt{n}}\gamma
\leq \frac{\mathcal{O}\left(T^{\frac{\alpha-1}{3\alpha-2}}\log^{\frac{1}{2}} 1/\delta \right)L + b}{\sqrt{n}}\gamma.
\end{align}
The bound in (\ref{eiglhelg}) also means that we have the following inequality uniformly for all $t = 1,...T$ with probability at least $1-3\delta$
\begin{align}\label{eiuyfiowy}\nonumber
&\| \nabla F(\mathbf{w}_{t+1}) -  \nabla F_S(\mathbf{w}_{t+1})\|^2\\&= \mathcal{O}\left(\frac{ T^{\frac{2\alpha-2}{3\alpha-2}}\log (\frac{1}{\delta}) }{n} \times \Big(d + \log (\frac{1}{\delta})\Big) \right).
\end{align}
It is clear that 
\begin{align}\label{epojvol}\nonumber
&\sum_{t=1}^T \| \nabla F(\mathbf{w}_{t}) \|^2\\\nonumber &\leq 2\sum_{t=1}^T\| \nabla F(\mathbf{w}_{t}) - \nabla F_S(\mathbf{w}_{t}) \|^2 + 2\sum_{t=1}^T\|  \nabla F_S(\mathbf{w}_{t}) \|^2\\\nonumber  &\leq 2T \max_{1 \leq t \leq T} \| \nabla F(\mathbf{w}_{t}) - \nabla F_S(\mathbf{w}_{t}) \|^2 + 2\sum_{t=1}^T \|  \nabla F_S(\mathbf{w}_{t}) \|^2\\ &= 2T  \| \nabla F(\mathbf{w}_{T}) - \nabla F_S(\mathbf{w}_{T}) \|^2 + 2\sum_{t=1}^T \|  \nabla F_S(\mathbf{w}_{t}) \|^2,
\end{align}
where the last equation follows from the fact that $\| \nabla F(\mathbf{w}_{t}) - \nabla F_S(\mathbf{w}_{t}) \|^2$ increases as the iterate number $t$ increases.

Then, plugging the bound in (\ref{eiuyfiowy}) and the bound in Theorem 5 into (\ref{epojvol}), we have the following inequality with probability at least $1-4\delta$
\begin{align*}
&\frac{1}{T} \sum_{t=1}^T \| \nabla F(\mathbf{w}_{t}) \|^2 
= \mathcal{O}\left(\frac{ T^{\frac{2\alpha-2}{3\alpha-2}}\log 1/\delta }{n} \times \Big(d + \log (\frac{1}{\delta})\Big) \right) \\&+ \mathcal{O}\left(\frac{\log 1/\delta}{T^\frac{2\alpha -2}{3\alpha - 2}}\right).
\end{align*}
Taking $T \asymp (\frac{n}{d})^{\frac{3\alpha-2}{4\alpha-4}}$, we have $\frac{1}{T} \sum_{t=1}^T \| \nabla F(\mathbf{w}_{t}) \|^2 = 
           \mathcal{O}\left(\Big(\frac{d}{n}\Big)^\frac{1}{2} \log 1/\delta \right)$ with probability at least $1 - 4\delta$,
which means that with probability at least $1 - \delta$
 \begin{align*}
\frac{1}{T} \sum_{t=1}^T \| \nabla F(\mathbf{w}_{t}) \|^2 & = 
           \mathcal{O}\left(\Big(\frac{d}{n}\Big)^\frac{1}{2} \log 1/\delta \right).
\end{align*}
The proof is complete.
\end{proof}
\subsection{Proof of Theorem 9}\label{apendixbb}
\begin{proof}
We consider two cases. 
Firstly, considering the case $\| \mathbf{m}_t \| \geq \tau_2$. Then we get
\begin{align*}
&F_S(\mathbf{w}_{t+1}) - F_S(\mathbf{w}_{t}) \\&\leq \langle \mathbf{w}_{t+1} - \mathbf{w}_{t}, \nabla F_S(\mathbf{w}_{t}) \rangle + \frac{L}{2}  \| \mathbf{w}_{t+1} -\mathbf{w}_{t}  \|^2\\
& \leq -\eta \langle  \bar{\mathbf{m}}_t, \nabla F_S(\mathbf{w}_{t}) \rangle + \frac{L}{2}\eta^2 \|\bar{\mathbf{m}}_t \|^2\\
&\leq -\eta \langle \frac{\tau_2\mathbf{m}_t}{\| \mathbf{m}_t \|} , \nabla F_S(\mathbf{w}_{t}) \rangle + \frac{L}{2}  \eta^2 \tau_2^2\\
&= \tau_2 \eta\Big(-\| \mathbf{m}_t\| +  \frac{\langle \mathbf{m}_t - \nabla F_S(\mathbf{w}_{t} ), \mathbf{m}_t \rangle}{\| \mathbf{m}_t \|} \Big) + \frac{L}{2}  \eta^2 \tau_2^2\\
&\leq \tau_2 \eta\Big(-\| \mathbf{m}_t \| + \| \mathbf{m}_t - \nabla F_S(\mathbf{w}_{t} )\| \Big) + \frac{L}{2}  \eta^2 \tau_2^2\\
&\leq \tau_2 \eta\Big(-\| \mathbf{m}_t \| + \| \mathbf{m}_t - \nabla F_S(\mathbf{w}_{t} )\| \\
&+ \frac{1}{3} (\| \mathbf{m}_t - \nabla F_S(\mathbf{w}_{t} ) \| + \| \mathbf{m}_t  \| -\| \nabla F_S(\mathbf{w}_{t} ) \|)\Big) + \frac{L}{2}  \eta^2 \tau_2^2\\
&\leq \eta\tau_2\Big(-\frac{1}{3}\| \nabla F_S(\mathbf{w}_{t})\| + \frac{4}{3}  \| \mathbf{m}_t - \nabla F_S(\mathbf{w}_{t} )\|\Big) + \frac{L}{2}  \eta^2 \tau_2^2.
\end{align*}
The above inequality implies that 
\begin{align*}
&\| \nabla F_S(\mathbf{w}_{t})\| \\&\leq  3\frac{F_S(\mathbf{w}_{t}) - F_S(\mathbf{w}_{t+1})}{\eta \tau_2}+4 \| \mathbf{m}_t - \nabla F_S(\mathbf{w}_{t}) \| + \frac{3L}{2}  \eta \tau_2.
\end{align*}
Thus, we get 
\begin{align}\label{777}\nonumber
&\frac{1}{T}\sum_{t=1}^T \| \nabla F_S(\mathbf{w}_{t})\| \leq 3\frac{F_S(\mathbf{w}_{1}) - F_S(\mathbf{w}_{T+1})}{\eta \tau_2 T} \\\nonumber&+  \frac{4}{T}\sum_{t=1}^T\| \mathbf{m}_t - \nabla F_S(\mathbf{w}_{t} )\| + \frac{3L}{2}  \eta \tau_2\\
&\leq 3\frac{F_S(\mathbf{w}_{1}) - F_S(\mathbf{w}_S^{\ast})}{\eta \tau_2 T}+  \frac{4}{T}\sum_{t=1}^T\| \mathbf{m}_t - \nabla F_S(\mathbf{w}_{t} )\| + \frac{3L}{2}  \eta \tau_2.
\end{align}
We now bound the term $\frac{1}{T}\sum_{t=1}^T\| \mathbf{m}_t - \nabla F_S(\mathbf{w}_{t} )\|$. Let $Z(a,b) = \nabla F_S(a) - \nabla F_S(b) $.
We have the following recursive formulation for any $t \geq 1$:
\begin{align*}
&\mathbf{m}_{t+1} \\&= \gamma (\mathbf{m}_{t}- \nabla F_S(\mathbf{w}_{t})+ \nabla F_S(\mathbf{w}_{t}) -\nabla F_S(\mathbf{w}_{t+1}) + \nabla F_S(\mathbf{w}_{t+1}) )\\& + (1-\gamma) (\nabla \bar{f}(\mathbf{w}_{t+1};z_{j_{t+1}})-\nabla F_S(\mathbf{w}_{t+1})+\nabla F_S(\mathbf{w}_{t+1}) )\\
&=\nabla F_S(\mathbf{w}_{t+1} ) + \gamma (\mathbf{m}_{t}- \nabla F_S(\mathbf{w}_{t})  + Z(\mathbf{w}_{t},\mathbf{w}_{t+1})) \\&+ (1-\gamma)(\nabla \bar{f}(\mathbf{w}_{t+1};z_{j_{t+1}})-\nabla F_S(\mathbf{w}_{t+1})).
\end{align*}
Then we get
\begin{align}\label{eqssss}\nonumber
&\mathbf{m}_{t+1} -\nabla F_S(\mathbf{w}_{t+1} )  = \gamma (\mathbf{m}_{t}- \nabla F_S(\mathbf{w}_{t})  + Z(\mathbf{w}_{t},\mathbf{w}_{t+1})) \\&+ (1-\gamma)(\nabla \bar{f}(\mathbf{w}_{t+1};z_{j_{t+1}})-\nabla F_S(\mathbf{w}_{t+1})).
\end{align}
For brevity,
define $\epsilon_t = \mathbf{m}_{t} - \nabla F_S(\mathbf{w}_{t})$ and $\epsilon'_t =\nabla \bar{f}(\mathbf{w}_t;z_{j_t})- \nabla F_S(\mathbf{w}_{t})$.
By setting $\mathbf{m}_{0}  = 0$ and $\mathbf{w}_{0} = \mathbf{w}_{1}$, we have $\epsilon_{0} = - \nabla F_S(\mathbf{w}_{1})$. Thus, the above recursion (\ref{eqssss}) of $\epsilon_{t+1}$ also holds for $t=0$.
We then unroll the recursion to get
\begin{align*}
&\epsilon_{t+1} =  (1-\gamma)\sum_{k = 0}^{t}\gamma^k \epsilon'_{t+1-k} \\&+ \gamma\sum_{k=0}^t \gamma^{k} Z(\mathbf{w}_{t-k},\mathbf{w}_{t-k+1}) +\gamma^{t+1} \epsilon_{0}.
\end{align*}
Clearly, we get
\begin{align*}
\|\epsilon_{t+1}\|& \leq  (1-\gamma)\|\sum_{k = 0}^{t}\gamma^k \epsilon'_{t+1-k}\|+ \gamma\sum_{k=0}^t \gamma^{k} L \eta \tau_2  +\gamma^{t+1} \|\epsilon_{0}\|\\
&\leq \gamma^{t+1} G + (1-\gamma)\|\sum_{k = 0}^{t}\gamma^k \epsilon'_{t+1-k}\|+ \frac{\gamma}{1-\gamma} L \eta \tau_2,
\end{align*}
where the first inequality holds due to the smoothness and the triangle inequality, and
where $G$ appears in the last inequality because of $\| \nabla F_S(\mathbf{w}_{1})\| \leq G$ (see (\ref{ineq00ji}) for details).

To give the bound of $\|  \epsilon_{t+1} \|$, we need to bound the term $\|\sum_{k = 0}^{t}\gamma^k \epsilon'_{t+1-k}\|$. We have
\begin{align*}
&\|\sum_{k = 0}^{t}\gamma^k \epsilon'_{t+1-k}\| \\&\leq \|\sum_{k = 0}^{t}\gamma^k \nabla \bar{f}(\mathbf{w}_{t+1-k};z_{j_{t+1-k}})- \nabla F_S(\mathbf{w}_{_{t+1-k}})\|\\
&\leq \|\sum_{k = 0}^{t}\gamma^k (\nabla \bar{f}(\mathbf{w}_{t+1-k};z_{j_{t+1-k}})- \mathbb{E}_{j_{t+1-k}}\nabla \bar{f}(\mathbf{w}_{t+1-k};z_{j_{t+1-k}}))\| \\& + \|\sum_{k = 0}^{t}\gamma^k ( \mathbb{E}_{j_{t+1-k}}\nabla \bar{f}(\mathbf{w}_{t+1-k};z_{j_{t+1-k}})- \nabla F_S(\mathbf{w}_{_{t+1-k}}))\|.
\end{align*}
Observing that $ \| \gamma^k (\nabla \bar{f}(\mathbf{w}_{t+1-k};z_{j_{t+1-k}})- \mathbb{E}_{j_{t+1-k}}\nabla \bar{f}(\mathbf{w}_{t+1-k};z_{j_{t+1-k}}))\| \leq 2\tau_1$ almost surely due to $\gamma < 1$ and according to (\ref{eqxuyao}), we can prove the following inequality with probability $1-\delta$ by Lemma 20,
\begin{align*}
&\|\sum_{k = 0}^{t}\gamma^k (\nabla \bar{f}(\mathbf{w}_{t+1-k};z_{j_{t+1-k}})- \mathbb{E}_{j_{t+1-k}}\nabla \bar{f}(\mathbf{w}_{t+1-k};z_{j_{t+1-k}}))\| \\
&\leq 2\left( \frac{2\tau_1}{3} + \Big(\sum_{k=0}^t \gamma^{2k} G^\alpha \tau_1^{2-\alpha} \Big)^{\frac{1}{2}} \right)\log \frac{2}{\delta}.
\end{align*}
And according to (\ref{eqyouxuyao}), there holds 
\begin{align*}
&\|\sum_{k = 0}^{t}\gamma^k ( \mathbb{E}_{j_{t+1-k}}\nabla \bar{f}(\mathbf{w}_{t+1-k};z_{j_{t+1-k}})- \nabla F_S(\mathbf{w}_{_{t+1-k}}))\| \\&\leq \sum_{k=0}^t\frac{\gamma^kG^\alpha}{\tau_1^{\alpha-1}}.
\end{align*}
Thus, with probability at least $1 - \delta$, we get
\begin{align*}
    &\|\sum_{k = 0}^{t}\gamma^k \epsilon'_{t+1-k}\|\\&\le \sum_{k=0}^t\frac{\gamma^kG^\alpha}{\tau_1^{\alpha-1}}+ 2\left( \frac{2\tau_1}{3} + \Big(\sum_{k=0}^t \gamma^{2k} G^\alpha \tau_1^{2-\alpha} \Big)^{\frac{1}{2}} \right)\log \frac{2}{\delta}\\
    &\leq \frac{4}{3}\tau_1\log(2/\delta) +\frac{G^\alpha}{(1-\gamma)\tau_1^{\alpha-1}}+ \frac{2\sqrt{ G^\alpha \tau_1^{2-\alpha}}}{(1-\gamma^2)^{1/2}}\log(2/\delta)\\
    &\leq \frac{4}{3}\tau_1\log(2/\delta) +\frac{G^\alpha}{(1-\gamma)\tau_1^{\alpha-1}}+ \frac{2\sqrt{ G^\alpha \tau_1^{2-\alpha}}}{(1-\gamma)^{1/2}}\log(2/\delta).
\end{align*}
Now, with probability at least $1 - \delta$, we get the bound of $\|  \epsilon_{t+1} \|$.
\begin{align*}
&\|\epsilon_{t+1}\| \leq \gamma^{t+1} G + (1-\gamma)\frac{4}{3}\tau_1\log(2/\delta) +\frac{G^\alpha}{\tau_1^{\alpha-1}}\\&+ 2\sqrt{ (1-\gamma)G^\alpha \tau_1^{2-\alpha}}\log(2/\delta)+ \frac{\gamma}{1-\gamma} L \eta \tau_2.
\end{align*}
This inequality means that with probability at least $1 - T\delta$
\begin{align*}
&\frac{1}{T}\sum_{t=1}^T\| \mathbf{m}_t - \nabla F_S(\mathbf{w}_{t} )\|\\&\leq \frac{1}{T}\sum_{t=1}^T \gamma^{t} G + (1-\gamma)\frac{4}{3}\tau_1\log(2/\delta) +\frac{G^\alpha}{\tau_1^{\alpha-1}}\\
&+ 2\sqrt{ (1-\gamma)G^\alpha \tau_1^{2-\alpha}}\log(2/\delta)+ \frac{\gamma}{1-\gamma} L \eta \tau_2\\
&\leq \frac{G}{T}\frac{\gamma}{1-\gamma} + (1-\gamma)\frac{4}{3}\tau_1\log(2/\delta) +\frac{G^\alpha}{\tau_1^{\alpha-1}}\\
&+ 2\sqrt{ (1-\gamma)G^\alpha \tau_1^{2-\alpha}}\log(2/\delta)+ \frac{\gamma}{1-\gamma} L \eta \tau_2.
\end{align*}
Choosing $\tau_1=\frac{pG}{(1-\gamma)^{1/\alpha}}$, where $p$ is a positive constant, we get
\begin{align*}
&\frac{1}{T}\sum_{t=1}^T\| \mathbf{m}_t - \nabla F_S(\mathbf{w}_{t} )\|\\&\leq \frac{G}{T}\frac{\gamma}{1-\gamma} + (1-\gamma)^{1-1/\alpha}\frac{4}{3}pG\log(2/\delta) \\&+Gp^{1-\alpha}(1-\gamma)^{1-1/\alpha}\\&+ 2Gp^{\frac{2-\alpha}{2}}(1-\gamma)^{1-1/\alpha}\log(2/\delta)+ \frac{\gamma}{1-\gamma} L \eta \tau_2\\
&\leq \frac{G}{T}\frac{\gamma}{1-\gamma} + \frac{\gamma}{1-\gamma} L \eta \tau_2 \\
&+ (1-\gamma)^{1-1/\alpha}\Big(\frac{4}{3}pG\log(2/\delta) + 2Gp^{\frac{2-\alpha}{2}}  \log(2/\delta) +  Gp^{1-\alpha}\Big).
\end{align*}
Next, setting $1-\gamma = \frac{s}{T^{\frac{\alpha}{3\alpha-2}}}$, $\eta = \frac{q}{T^{\frac{\alpha}{3\alpha-2}}}$, and $\tau_2 = \frac{r}{T^{\frac{\alpha-1}{3\alpha-2}}}$, where $s$, $q$, and $r$ are positive constants such that $1-\gamma \leq 1$, we get 
\begin{align}\label{eq666}\nonumber
&\frac{1}{T}\sum_{t=1}^T\| \mathbf{m}_t - \nabla F_S(\mathbf{w}_{t} )\|\\&\leq \mathcal{O}\left(\frac{1}{T^{\frac{2\alpha-2}{3\alpha-2}}} + \frac{1}{T^{\frac{\alpha-1}{3\alpha-2}}}\log(1/\delta) + \frac{1}{T^{\frac{\alpha-1}{3\alpha-2}}} \right).
\end{align}
Plugging the bound in (\ref{eq666}) into (\ref{777}), we get the following inequality with probability at least $1 - T\delta$
\begin{align*}
&\frac{1}{T}\sum_{t=1}^T \| \nabla F_S(\mathbf{w}_{t})\| \\&\leq \mathcal{O}\left(\frac{1}{T^{\frac{\alpha-1}{3\alpha-2}}} + \frac{1}{T^{\frac{2\alpha-2}{3\alpha-2}}} + \frac{1}{T^{\frac{\alpha-1}{3\alpha-2}}}\log(\frac{1}{\delta}) + \frac{1}{T^{\frac{\alpha-1}{3\alpha-2}}}+   \frac{1}{T^{\frac{2\alpha -1}{3\alpha-2}}} \right)\\&=\mathcal{O}\left(\frac{\log(\frac{1}{\delta})}{T^{\frac{\alpha-1}{3\alpha-2}}} \right),
\end{align*}
which means that we get with probability at least $1 - \delta$
\begin{align*}
&\frac{1}{T}\sum_{t=1}^T \| \nabla F_S(\mathbf{w}_{t})\| \leq \mathcal{O}\left( \frac{1}{T^{\frac{\alpha-1}{3\alpha-2}}} \log(T/\delta) \right).
\end{align*}

We then consider the case $\| \mathbf{m}_t \| < \tau_2$, we get 
\begin{align*}
&F_S(\mathbf{w}_{t+1}) - F_S(\mathbf{w}_{t}) \leq  -\eta \langle  \bar{\mathbf{m}}_t, \nabla F_S(\mathbf{w}_{t}) \rangle + \frac{1}{2} L \eta^2 \|\bar{\mathbf{m}}_t \|^2\\
&\leq -\eta \langle \mathbf{m}_t, \nabla F_S(\mathbf{w}_{t}) \rangle + \frac{1}{2} L \eta^2 \tau_2^2\\
&= -\eta \langle \mathbf{m}_t - \nabla F_S(\mathbf{w}_{t}) + \nabla F_S(\mathbf{w}_{t}), \nabla F_S(\mathbf{w}_{t})  \rangle + \frac{1}{2} L \eta^2 \tau_2^2\\
&= -\eta \langle \mathbf{m}_t-\nabla F_S(\mathbf{w}_{t}), \nabla F_S(\mathbf{w}_{t}) \rangle - \eta \| \nabla F_S(\mathbf{w}_{t})\|^2 + \frac{1}{2} L \eta^2 \tau_2^2\\
&\leq \frac{\eta}{2} \|\mathbf{m}_t-\nabla F_S(\mathbf{w}_{t})\|^2 +\frac{\eta}{2} \| \nabla F_S(\mathbf{w}_{t})\|^2 \\
&- \eta \| \nabla F_S(\mathbf{w}_{t})\|^2 + \frac{1}{2} L \eta^2 \tau_2^2,
\end{align*}
which implies that
\begin{align*}
&\| \nabla F_S(\mathbf{w}_{t} ) \|^2 \leq   \frac{2(F_S(\mathbf{w}_{t}) - F_S(\mathbf{w}_{t+1}))}{\eta} \\
&+ \| \nabla F_S(\mathbf{w}_{t}) - \mathbf{m}_t\|^2  +  L \eta \tau_2^2.
\end{align*}
Till here, we can derive that
\begin{align}\label{eq777}\nonumber
&\frac{1}{T}\sum_{t=1}^T \| \nabla F_S(\mathbf{w}_{t})\|^2 \leq  \frac{2(F_S(\mathbf{w}_{1}) - F_S(\mathbf{w}_S^{\ast}))}{ T\eta}+  L \eta \tau_2^2 \\
&+ \frac{1}{T}\sum_{t=1}^T\| \mathbf{m}_t - \nabla F_S(\mathbf{w}_{t}) \|^2.
\end{align}
Recall that $\epsilon_t = \mathbf{m}_{t} - \nabla F_S(\mathbf{w}_{t})$. We have known that with probability at least $1 - \delta$ 
\begin{align*}
&\|\epsilon_{t+1}\| \leq \gamma^{t+1} G + (1-\gamma)\frac{4}{3}\tau_1\log(2/\delta) +\frac{G^\alpha}{\tau_1^{\alpha-1}}\\&+ 2\sqrt{ (1-\gamma)G^\alpha \tau_1^{2-\alpha}}\log(2/\delta)+ \frac{\gamma}{1-\gamma} L \eta \tau_2.
\end{align*}
Thus, we get 
\begin{align*}
&\|\epsilon_{t+1}\|^2 \leq 4(\gamma^{2t+2} G^2 + (1-\gamma)^2\tau_1^2\log^2(2/\delta) +\frac{G^{2\alpha}}{\tau_1^{2\alpha-2}}\\&+ (1-\gamma)G^\alpha \tau_1^{2-\alpha}\log^2(2/\delta)+\frac{\gamma^2}{(1-\gamma)^2} L^2 \eta^2 \tau_2^2).
\end{align*}

Besides, we have proved that when $\tau_1=\frac{pG}{(1-\gamma)^{1/\alpha}}$, $1-\gamma = \frac{s}{T^{\frac{\alpha}{3\alpha-2}}}$, $\eta = \frac{q}{T^{\frac{\alpha }{3\alpha-2}}}$, and $\tau_2 = \frac{r}{T^{\frac{\alpha-1}{3\alpha-2}}}$,
\begin{align*}
&\frac{1}{T}\sum_{t=1}^T\| \mathbf{m}_t - \nabla F_S(\mathbf{w}_{t} )\| \\&\leq \mathcal{O}\left(\frac{1}{T^{\frac{2\alpha-2}{3\alpha-2}}} + \frac{1}{T^{\frac{\alpha-1}{3\alpha-2}}} \log(1/\delta)+ \frac{1}{T^{\frac{\alpha-1}{3\alpha-2}}} \right).
\end{align*}
With the similar proof pattern, we get the following inequality with probability at least $1 - T\delta$ 
\begin{align*}
&\frac{1}{T}\sum_{t=1}^T\| \mathbf{m}_t - \nabla F_S(\mathbf{w}_{t}) \|^2 \\& \leq  4(\frac{\gamma^2G^2}{T(1-\gamma^2)}  + (1-\gamma)^{2-\frac{2}{\alpha}}(p^2G^2\log^2(2/\delta)+ G^2p^{2-2\alpha } \\&+ G^\alpha p^{2-\alpha} G^{2-\alpha}\log^2 (2/\delta))+\frac{\gamma^2}{(1-\gamma)^2} L^2 \eta^2 \tau_2^2)\\
&\leq \mathcal{O}\left(\frac{1}{T^{\frac{2\alpha-2}{3\alpha-2}}} + \frac{1}{T^{\frac{2\alpha-2}{3\alpha-2}}}\log^2(1/\delta) + \frac{1}{T^{\frac{2\alpha-2}{3\alpha-2}}}\right)\\&= \mathcal{O}\left(\frac{1}{T^{\frac{2\alpha-2}{3\alpha-2}}} \log^2(1/\delta) \right).
\end{align*}
Plugging this bound into (\ref{eq777}), with probability at least $1 - T\delta$ we have
\begin{align*}
&\frac{1}{T}\sum_{t=1}^T \| \nabla F_S(\mathbf{w}_{t})\|^2 \\&\leq \mathcal{O}\left(\frac{1}{T^{\frac{2\alpha-2}{3\alpha-2}}} + \frac{1}{T^{\frac{3\alpha-2}{3\alpha-2}}} + \frac{1}{T^{\frac{2\alpha-2}{3\alpha-2}}}\log^2(1/\delta) \right),
\end{align*}
which means that 
with probability at least $1 - \delta$ we have
\begin{align*}
&\frac{1}{T}\sum_{t=1}^T \| \nabla F_S(\mathbf{w}_{t})\|^2 \leq \mathcal{O}\left(\frac{1}{T^{\frac{2\alpha-2}{3\alpha-2}}}\log^2(T/\delta)  \right).
\end{align*}
According to the Jensen's inequality, we have
\begin{align*}
\left( \frac{1}{T}\sum_{t=1}^T \| \nabla F_S(\mathbf{w}_{t})\|\right)^2 \leq \frac{1}{T}\sum_{t=1}^T \| \nabla F_S(\mathbf{w}_{t})\|^2.
\end{align*}
Hence, we finally get with probability at least $1 - \delta$
\begin{align*}
&\frac{1}{T}\sum_{t=1}^T \| \nabla F_S(\mathbf{w}_{t})\| \leq \mathcal{O}\left( \frac{1}{T^{\frac{\alpha-1}{3\alpha-2}}} \log(T/\delta) \right).
\end{align*}

Combining the two cases, we can conclude that with probability at least $1 - \delta$
\begin{align*}
&\frac{1}{T}\sum_{t=1}^T \| \nabla F_S(\mathbf{w}_{t})\| \leq \mathcal{O}\left( \frac{1}{T^{\frac{\alpha-1}{3\alpha-2}}} \log \frac{T}{\delta} \right).
\end{align*}
The proof is complete.
\end{proof}

\subsection{Proof of Theorem 11}

\begin{proof}
Similar to the proof of Theorem 7, we have the following inequality with probability $1-\delta$
\begin{align*}
&\frac{1}{T}\sum_{t=1}^T \| \nabla F(\mathbf{w}_{t}) \| \\  \leq &\frac{1}{T}\sum_{t=1}^T \max_{ t \leq T} \| \nabla F(\mathbf{w}_{t}) - \nabla F_S(\mathbf{w}_{t}) \| + \frac{1}{T}\sum_{t=1}^T \|  \nabla F_S(\mathbf{w}_{t}) \|\\\leq & \max_{ t \leq T} \| \nabla F(\mathbf{w}_{t}) - \nabla F_S(\mathbf{w}_{t}) \| +  \mathcal{O}\left( \frac{1}{T^{\frac{\alpha-1}{3\alpha-2}}} \log \frac{T}{\delta} \right),
\end{align*}
where the last inequality holds due to Theorem 9.
Moreover, we know that
$\mathbf{w}_{t+1} = \mathbf{w}_t - \eta_t \bar{\mathbf{m}}_t$.
Then, using $\mathbf{w}_1 = 0$, it is clear that
\begin{align*}
\mathbf{w}_{t+1} = - \sum_{i=1}^t \eta_i \bar{\mathbf{m}}_i.
\end{align*}
Further, we have
\begin{align}\label{eqjijiji}
\|\mathbf{w}_{t+1}\| \leq  \sum_{i=1}^t \eta_i \tau_2 = t \eta \tau_2.
\end{align}
Since $\eta = \frac{q}{T^{\frac{\alpha}{3\alpha-2}}}$ and $\tau_2 = \frac{r}{T^{\frac{\alpha-1}{3\alpha-2}}}$, we get $\|\mathbf{w}_{t+1}\| =  \frac{qr t}{T^{\frac{2\alpha -1}{3\alpha-2} }}$. For brevity, denoted by $\gamma = (2+ 2\sqrt{48e\sqrt{2}(\log 2 + d \log(3e))} + \sqrt{2 \log (\frac{1}{\delta})})$. According to Lemma 21,  with probability $1-\delta$ we have 
\begin{align*}
&\max_{ t \leq T} \| \nabla F(\mathbf{w}_{t}) - \nabla F_S(\mathbf{w}_{t}) \| \\&\leq \frac{(L R_{T} + b)\gamma}{\sqrt{n}}
= \frac{(L \|\mathbf{w}_{T}\| + b)\gamma}{\sqrt{n}}
\leq \frac{(L \frac{qr T}{T^{\frac{2\alpha -1}{3\alpha-2}} } + b)\gamma}{\sqrt{n}},
\end{align*}
where the last inequality holds due to (\ref{eqjijiji}).
Next, we can derive the following inequality with probability $1-2\delta$
\begin{align*}
&\frac{1}{T}\sum_{t=1}^T \| \nabla F(\mathbf{w}_{t}) \| \\&=\mathcal{O}\left(\frac{ T^{\frac{\alpha-1}{3\alpha-2}}}{\sqrt{n}} \times \Big(\sqrt{d} + \log^{\frac{1}{2}} (\frac{1}{\delta})\Big) \right) +   \mathcal{O}\left( \frac{1}{T^{\frac{\alpha-1}{3\alpha-2}}} \log \frac{T}{\delta} \right).
\end{align*}


Taking $T \asymp (\frac{n}{d})^{\frac{3\alpha-2}{4\alpha-4}}$, we have $\frac{1}{T} \sum_{t=1}^T \| \nabla F(\mathbf{w}_{t}) \|  = 
            \mathcal{O} \left( \Big(\frac{d}{n}\Big)^{\frac{1}{4}} \log\frac{n}{d\delta} \right)$ with probability $1 - 2\delta$,
which means 
with probability at least $1 - \delta$ we have
 \begin{align*}
\frac{1}{T} \sum_{t=1}^T \| \nabla F(\mathbf{w}_{t}) \| & = 
            \mathcal{O} \left( \Big(\frac{d}{n}\Big)^{\frac{1}{4}} \log\frac{n}{d\delta} \right).
\end{align*}
The proof is complete.
\end{proof}
\subsection{Proof of Theorem 13}

\begin{proof}
With the descent lemma of smoothness, we have
\begin{align*}
&F_S(\mathbf{w}_{t+1}) - F_S(\mathbf{w}_{t}) \\&\leq \langle \mathbf{w}_{t+1} - \mathbf{w}_{t}, \nabla F_S(\mathbf{w}_{t}) \rangle + \frac{1}{2} L \| \mathbf{w}_{t+1} -\mathbf{w}_{t}  \|^2\\
& = -\eta_t \langle  \nabla \bar{f}(\mathbf{w}_t;z_{j_t}) - \nabla F_S(\mathbf{w}_{t}), \nabla F_S(\mathbf{w}_{t}) \rangle -\eta_t \| \nabla F_S(\mathbf{w}_{t}) \|^2 \\&+ \frac{1}{2} L \eta_t^2 \|\nabla \bar{f}(\mathbf{w}_t;z_{j_t}) \|^2 ,
\end{align*}
which implies that 
\begin{align}\label{thelastinequali}\nonumber
&\sum_{t=1}^T\| \nabla F_S(\mathbf{w}_{t}) \|^2 \leq \sum_{t=1}^T\frac{ F_S(\mathbf{w}_{t}) - F_S(\mathbf{w}_{t+1}) }{\eta_t} \\\nonumber&- \sum_{t=1}^T\langle  \nabla \bar{f}(\mathbf{w}_t;z_{j_t}) - \nabla F_S(\mathbf{w}_{t}), \nabla F_S(\mathbf{w}_{t}) \rangle \\\nonumber& + \frac{1}{2} L \sum_{t=1}^T\eta_t  \|\nabla \bar{f}(\mathbf{w}_t;z_{j_t}) \|^2\\\nonumber
&= \sum_{t=1}^T\Big(\frac{ F_S(\mathbf{w}_{t}) - F_S(\mathbf{w}_{S}^{\ast}) }{\eta_t}- \frac{ F_S(\mathbf{w}_{t+1}) - F_S(\mathbf{w}_{S}^{\ast}) }{\eta_t}\Big ) \\\nonumber&- \sum_{t=1}^T\langle  \nabla \bar{f}(\mathbf{w}_t;z_{j_t}) - \nabla F_S(\mathbf{w}_{t}), \nabla F_S(\mathbf{w}_{t}) \rangle \\\nonumber
&+ \frac{1}{2} L \sum_{t=1}^T\eta_t  \|\nabla \bar{f}(\mathbf{w}_t;z_{j_t}) \|^2\\\nonumber
&\leq \frac{F_S(\mathbf{w}_{1}) - F_S(\mathbf{w}_{S}^{\ast})}{\eta_1}+\sum_{t=2}^T(\frac{ 1 }{\eta_t}-\frac{ 1 }{\eta_{t-1}}) (F_S(\mathbf{w}_{t}) - F_S(\mathbf{w}_{S}^{\ast}))  \\\nonumber
&- \sum_{t=1}^T\langle  \nabla \bar{f}(\mathbf{w}_t;z_{j_t}) - \nabla F_S(\mathbf{w}_{t}), \nabla F_S(\mathbf{w}_{t}) \rangle \\\nonumber&+ \frac{1}{2} L \sum_{t=1}^T\eta_t  \|\nabla \bar{f}(\mathbf{w}_t;z_{j_t}) \|^2\\\nonumber
&\leq \frac{2M}{\eta_T}- \sum_{t=1}^T\langle  \nabla \bar{f}(\mathbf{w}_t;z_{j_t}) - \nabla F_S(\mathbf{w}_{t}), \nabla F_S(\mathbf{w}_{t}) \rangle \\\nonumber&+ \frac{1}{2} L \sum_{t=1}^T\eta_t  \|\nabla \bar{f}(\mathbf{w}_t;z_{j_t}) \|^2\\\nonumber
&\leq  2M\sqrt{G_0^2 + \sum_{t=1}^T \|\nabla \bar{f}(\mathbf{w}_t;z_{j_t})\|^2}  \\\nonumber&- \sum_{t=1}^T \langle  \nabla \bar{f}(\mathbf{w}_t;z_{j_t}) - \nabla F_S(\mathbf{w}_{t}), \nabla F_S(\mathbf{w}_{t}) \rangle  \\\nonumber&+  L \sqrt{G_0^2 + \sum_{t=1}^T \|\nabla \bar{f}(\mathbf{w}_t;z_{j_t})\|^2}  \\\nonumber
& = (2M + L)\sqrt{G_0^2 + \sum_{t=1}^T \|\nabla \bar{f}(\mathbf{w}_t;z_{j_t})\|^2} \\&- \sum_{t=1}^T \langle  \nabla \bar{f}(\mathbf{w}_t;z_{j_t}) - \nabla F_S(\mathbf{w}_{t}), \nabla F_S(\mathbf{w}_{t}) \rangle,
\end{align}
where the last inequality holds due to the definition of $\eta_t$ and Lemma 22.

We now bound the term $- \sum_{t=1}^T \langle  \nabla \bar{f}(\mathbf{w}_t;z_{j_t}) - \nabla F_S(\mathbf{w}_{t}), \nabla F_S(\mathbf{w}_{t}) \rangle$. We have
\begin{align*}
&  - \sum_{t=1}^T \langle  \nabla \bar{f}(\mathbf{w}_t;z_{j_t}) - \nabla F_S(\mathbf{w}_{t}), \nabla F_S(\mathbf{w}_{t}) \rangle\\
&=- \sum_{t=1}^T \langle  \nabla \bar{f}(\mathbf{w}_t;z_{j_t}) - \mathbb{E}_{j_t}\nabla \bar{f}(\mathbf{w}_t;z_{j_t}) \\&+ \mathbb{E}_{j_t}\nabla \bar{f}(\mathbf{w}_t;z_{j_t}) - \nabla F_S(\mathbf{w}_{t}), \nabla F_S(\mathbf{w}_{t}) \rangle\\
&=- \sum_{t=1}^T \langle  \nabla \bar{f}(\mathbf{w}_t;z_{j_t}) - \mathbb{E}_{j_t}\nabla \bar{f}(\mathbf{w}_t;z_{j_t}) , \nabla F_S(\mathbf{w}_{t}) \rangle \\&- \sum_{t=1}^T \langle\mathbb{E}_{j_t}\nabla \bar{f}(\mathbf{w}_t;z_{j_t}) - \nabla F_S(\mathbf{w}_{t}), \nabla F_S(\mathbf{w}_{t}) \rangle.
\end{align*}
Since $ \mathbb{E}_{j_t}[  \langle \nabla \bar{f}(\mathbf{w}_t;z_{j_t}) - \mathbb{E}_{j_t}\nabla \bar{f}(\mathbf{w}_t;z_{j_t}), \nabla F_S(\mathbf{w}_{t}) \rangle ]= 0$, thus the sequence $\{-   \langle \nabla \bar{f}(\mathbf{w}_t;z_{j_t}) - \mathbb{E}_{j_t}\nabla \bar{f}(\mathbf{w}_t;z_{j_t}), \nabla F_S(\mathbf{w}_{t}) \rangle, t \in \mathbb{N} \}$ is a martingale difference sequence. Denoted by $\xi_t = -   \langle \nabla \bar{f}(\mathbf{w}_t;z_{j_t}) - \mathbb{E}_{j_t}\nabla \bar{f}(\mathbf{w}_t;z_{j_t}), \nabla F_S(\mathbf{w}_{t}) \rangle$. 
There holds
\begin{align*}
|\xi_t|\leq   (\|  \nabla \bar{f}(\mathbf{w}_t;z_{j_t}) \| + \| \mathbb{E}_{j_t}\nabla \bar{f}(\mathbf{w}_t;z_{j_t})\| )\| \nabla F_S(\mathbf{w}_{t}) \| \leq 2 \tau G,
\end{align*}
and 
\begin{align*}
 &\mathbb{E}_{j_t} [\| \nabla \bar{f}(\mathbf{w}_t;z_{j_t}) - \mathbb{E}_{j_t}\nabla \bar{f}(\mathbf{w}_t;z_{j_t})\|^2 ] \leq  \mathbb{E}_{j_t} [\| \nabla \bar{f}(\mathbf{w}_t;z_{j_t}) \|^2 ] \\&\leq \mathbb{E}_{j_t} [\| \nabla f(\mathbf{w}_t;z_{j_t}) \|^\alpha \tau^{2-\alpha}] \leq G^\alpha \tau^{2-\alpha}.
\end{align*}
And according to (\ref{eqzuihouyigeshizi}), we have
\begin{align*}
&\sum_{t=1}^T  \mathbb{E}_{j_t} [(\xi_t - \mathbb{E}_{j_t} \xi_t)^2] \\&\leq \sum_{t=1}^T \mathbb{E}_{j_t} [\| \nabla \bar{f}(\mathbf{w}_t;z_{j_t}) - \mathbb{E}_{j_t}\nabla \bar{f}(\mathbf{w}_t;z_{j_t})\|^2 \| \nabla F_S(\mathbf{w}_{t}) \|^2] \\&\leq G^\alpha \tau^{2-\alpha} \sum_{t=1}^T \|\nabla F_S(\mathbf{w}_{t}) \|^2.
\end{align*}
Thus by Lemma 19, with probability $1- \delta$, we have
\begin{align*}
\sum_{t=1}^T \xi_t \leq \frac{\rho G^\alpha \tau^{2-\alpha} \sum_{t=1}^T   \|\nabla F_S(\mathbf{w}_{t}) \|^2}{2\tau G} + \frac{2 \tau G \log(1/\delta)}{\rho}.
\end{align*}
Taking $\rho = \min\left\{\frac{G^{1-\alpha}}{2\tau^{1-\alpha}},1\right\} $, we get 
\begin{align}\label{danxinjiayi}
\sum_{t=1}^T \xi_t \leq \frac{\sum_{t=1}^T  \|\nabla F_S(\mathbf{w}_{t}) \|^2}{4} + \max\{4  \tau^{2-\alpha} G^{\alpha}, 2  \tau G\} \log(1/\delta).
\end{align}
Furthermore, we bound the term $-  \sum_{t=1}^T \langle\mathbb{E}_{j_t}\nabla \bar{f}(\mathbf{w}_t;z_{j_t}) -  \nabla F_S(\mathbf{w}_{t})  , \nabla F_S(\mathbf{w}_{t}) \rangle$. Similar to (\ref{eq999}), we have
\begin{align}\label{youdiandanxin}\nonumber
&-  \sum_{t=1}^T \langle\mathbb{E}_{j_t}\nabla \bar{f}(\mathbf{w}_t;z_{j_t}) -  \nabla F_S(\mathbf{w}_{t})  , \nabla F_S(\mathbf{w}_{t}) \rangle\\& \leq \frac{1}{2}G^{2\alpha} \tau^{2-2\alpha}T  + \frac{1}{2}\sum_{t=1}^T   \| \nabla F_S(\mathbf{w}_{t}) \|^2.
\end{align}
Thus, with probability $1- \delta$, we now get
\begin{align*}
&\frac{1}{4}\sum_{t=1}^T \| \nabla F_S(\mathbf{w}_{t}) \|^2  \leq  (2M + L)\sqrt{G_0^2 + \sum_{t=1}^T \|\nabla \bar{f}(\mathbf{w}_t;z_{j_t})\|^2} \\
&+ \frac{1}{2}G^{2\alpha} \tau^{2-2\alpha}T +   \max\{4  \tau^{2-\alpha} G^{\alpha}, 2  \tau G\} \log(1/\delta).
\end{align*}
Since $\tau  = p T^{\frac{1}{3\alpha - 2}}$ for some positive constant $p$, the above bound implies
\begin{align}\label{eq0078}\nonumber
&\frac{1}{4}\sum_{t=1}^T \| \nabla F_S(\mathbf{w}_{t}) \|^2 \leq  (2M + L)\sqrt{G_0^2 + \sum_{t=1}^T \|\nabla \bar{f}(\mathbf{w}_t;z_{j_t})\|^2} \\
& + \mathcal{O}\left(T^{\frac{2-2\alpha}{3\alpha-2}}T \log(1/\delta)\right).
\end{align}
We now bound the term $\sum_{t=1}^T \|\nabla \bar{f}(\mathbf{w}_t;z_{j_t})\|^2$.
\begin{align*}
&\sum_{t=1}^T \|\nabla \bar{f}(\mathbf{w}_t;z_{j_t})\|^2 \\&\leq 
3\sum_{t=1}^T \Big(\|\nabla \bar{f}(\mathbf{w}_t;z_{j_t}) - \mathbb{E}_{j_t} \nabla \bar{f}(\mathbf{w}_t;z_{j_t})\|^2 \\
& - \mathbb{E}_{j_t} \|\nabla \bar{f}(\mathbf{w}_t;z_{j_t}) - \mathbb{E}_{j_t} \nabla \bar{f}(\mathbf{w}_t;z_{j_t})\|^2 \\
&+  \mathbb{E}_{j_t} \|\nabla \bar{f}(\mathbf{w}_t;z_{j_t}) - \mathbb{E}_{j_t} \nabla \bar{f}(\mathbf{w}_t;z_{j_t})\|^2 \\
& + \|\mathbb{E}_{j_t} \nabla \bar{f}(\mathbf{w}_t;z_{j_t}) - \nabla F_S(\mathbf{w}_{t}) \|^2 + \| \nabla F_S(\mathbf{w}_{t}) \|^2\Big).
\end{align*}
From the proof of Theorem 5, we know that $\| \mathbb{E}_{j_t}\nabla \bar{f}(\mathbf{w}_t;z_{j_t}) -  \nabla F_S(\mathbf{w}_{t})  \| \leq G^\alpha \tau^{1-\alpha}$, $\sum_{t=1}^T \mathbb{E}_{j_t} \|\nabla \bar{f}(\mathbf{w}_t;z_{j_t}) - \mathbb{E}_{j_t}\nabla \bar{f}(\mathbf{w}_t;z_{j_t})\|^2\leq   TG^\alpha\tau^{2-\alpha}$, and that when $\tau \leq  T^{\frac{1}{\alpha}}$, i.e., $p\leq T^{\frac{2\alpha-2}{\alpha(3\alpha-2)}}$,
with probability at least $1-\delta$ there holds
$\sum_{t=1}^T \|\nabla \bar{f}(\mathbf{w}_t;z_{j_t}) - \mathbb{E}_{j_t}\nabla \bar{f}(\mathbf{w}_t;z_{j_t})\|^2-\mathbb{E}_{j_t}\|\nabla \bar{f}(\mathbf{w}_t;z_{j_t}) - \mathbb{E}_{j_t}\nabla \bar{f}(\mathbf{w}_t;z_{j_t})\|^2 \leq \sqrt{T}\tau^{2-\frac{\alpha}{2}} G^\alpha +4 \sqrt{T}\tau^{2-\frac{\alpha}{2}}\log(1/\delta)$.

Thus, with probability at least $1-\delta$, we have
\begin{align}\label{eq007}\nonumber
&\sum_{t=1}^T \|\nabla \bar{f}(\mathbf{w}_t;z_{j_t})\|^2 \\\nonumber&\leq 
3G^{2\alpha} \tau^{2-2\alpha}T+  3TG^\alpha\tau^{2-\alpha}+ 3\sqrt{T}\tau^{2-\frac{\alpha}{2}} G^\alpha \\\nonumber
& +12 \sqrt{T}\tau^{2-\frac{\alpha}{2}}\log(1/\delta) +3\sum_{t=1}^T \| \nabla F_S(\mathbf{w}_{t}) \|^2\\&
\leq \mathcal{O}(TT^{\frac{2-\alpha}{3\alpha-2}}\log(1/\delta)) +3\sum_{t=1}^T \| \nabla F_S(\mathbf{w}_{t}) \|^2.
\end{align}
Plugging (\ref{eq007}) into (\ref{eq0078}), we get with probability at least $1-2\delta$
\begin{align}\label{eq00789}\nonumber
&\frac{1}{4}\sum_{t=1}^T \| \nabla F_S(\mathbf{w}_{t}) \|^2  \leq  (2M + L)\\\nonumber
&\times\sqrt{G_0^2 + \mathcal{O}(TT^{\frac{2-\alpha}{3\alpha-2}}\log(1/\delta)) +3\sum_{t=1}^T \| \nabla F_S(\mathbf{w}_{t}) \|^2} \\
& + \mathcal{O}(T^{\frac{2-2\alpha}{3\alpha-2}}T \log(1/\delta)).
\end{align}
Then, solving the quadratic inequality of $\sum_{t=1}^T \| \nabla F_S(\mathbf{w}_{t}) \|^2$, we get the following inequality with probability at least $1-2\delta$
\begin{align*}
\sum_{t=1}^T \| \nabla F_S(\mathbf{w}_{t}) \|^2 \leq \mathcal{O}\left(T^{\frac{\alpha}{3\alpha-2}} \log(1/\delta) \right).
\end{align*}
Thus, we finally obtain the following inequality with probability $1-\delta$
\begin{align*}
\frac{1}{T}\sum_{t=1}^T \| \nabla F_S(\mathbf{w}_{t}) \|^2 
\leq \mathcal{O}\left(\frac{\log 1/\delta}{T^\frac{2\alpha -2}{3\alpha - 2}} \right).
\end{align*}
The proof is complete.
\end{proof}
\subsection{Proof of Theorem 15}
\begin{proof}
Firstly, we have the following inequality with probability $1-\delta$
\begin{align*}
&\frac{1}{T}\sum_{t=1}^T \| \nabla F(\mathbf{w}_{t}) \|^2 \\\leq &\frac{2}{T}\sum_{t=1}^T \| \nabla F(\mathbf{w}_{t}) - \nabla F_S(\mathbf{w}_{t}) \|^2 + \frac{2}{T}\sum_{t=1}^T \|  \nabla F_S(\mathbf{w}_{t}) \|^2\\  \leq &\frac{2}{T}\sum_{t=1}^T \max_{ t \leq T} \| \nabla F(\mathbf{w}_{t}) - \nabla F_S(\mathbf{w}_{t}) \|^2 + \frac{2}{T}\sum_{t=1}^T \|  \nabla F_S(\mathbf{w}_{t}) \|^2 \\\leq & 2\max_{ t \leq T} \| \nabla F(\mathbf{w}_{t}) - \nabla F_S(\mathbf{w}_{t}) \|^2 +  \mathcal{O}\left( \frac{1}{T^{\frac{2\alpha-2}{3\alpha-2}}} \log\frac{1}{\delta} \right),
\end{align*}
where the last inequality follows from the result of Theorem 13.

Moreover, we know that
\begin{align*}
\mathbf{w}_{t+1} = \mathbf{w}_t - \eta_t \nabla \bar{f}(\mathbf{w}_t;z_{j_t}).
\end{align*}
Then, using $\mathbf{w}_{1}=0$, it is clear that
\begin{align*}
\mathbf{w}_{t+1} = - \sum_{i=1}^t \eta_i \nabla \bar{f}(\mathbf{w}_i;z_{j_i}).
\end{align*}
Further, we have
\begin{align*}
\|\mathbf{w}_{t+1}\| =  \|\sum_{i=1}^t \eta_i\nabla \bar{f}(\mathbf{w}_i;z_{j_i})\|.
\end{align*}
Thus, 
\begin{align*}
&\|\mathbf{w}_{T+1}\|^2 =  \|\sum_{t=1}^{T} \eta_t\nabla \bar{f}(\mathbf{w}_t;z_{j_t})\|^2 \leq (\sum_{t=1}^{T}\| \eta_t\nabla \bar{f}(\mathbf{w}_t;z_{j_t})\|)^2\\& \leq T\sum_{t=1}^{T} \|\eta_t \nabla \bar{f}(\mathbf{w}_t;z_{j_t})\|^2  = T\sum_{t=1}^{T} \frac{\|\nabla \bar{f}(\mathbf{w}_t;z_{j_t})\|^2}{G_0^2 + \sum_{i=1}^t \| \nabla \bar{f}(\mathbf{w}_i;z_{j_i})\|^2}\\& \leq T+ T\log \left(1+\sum_{t=1}^{T} \| \nabla \bar{f}(\mathbf{w}_t;z_{j_t})\|^2 \right) \leq T+ T\log (1+T\tau^2)\\&\leq \mathcal{O}\left(T\log (1+T^{\frac{3\alpha}{3\alpha-2}})\right),
\end{align*}
where the third inequality follows from Lemma
22, and where the last inequality follows from that $\tau  = p T^{\frac{1}{3\alpha - 2}}$.
For brevity, denoted by $\gamma = (2+ 2\sqrt{48e\sqrt{2}(\log 2 + d \log(3e))} + \sqrt{2 \log (\frac{1}{\delta})})$. According to Lemma 21,  with probability $1-\delta$ we have 
\begin{align*}
&\max_{ t \leq T} \| \nabla F(\mathbf{w}_{t}) - \nabla F_S(\mathbf{w}_{t}) \|^2 \leq \frac{(L R_{T} + b)^2\gamma^2}{n}
\\&= \frac{(L \|\mathbf{w}_{T}\| + b)^2\gamma^2}{\sqrt{n}}\leq \mathcal{O}\left(\frac{ T \log (1+T^{\frac{3\alpha}{3\alpha-2}})}{n} \times \Big(d+ \log (\frac{1}{\delta})\Big) \right).
\end{align*}
Next, we can derive the following inequality with probability $1-2\delta$
\begin{align*}
&\frac{1}{T}\sum_{t=1}^T \| \nabla F(\mathbf{w}_{t}) \|^2 \\&=\mathcal{O}\left(\frac{ T \log (1+T^{\frac{3\alpha}{3\alpha-2}})}{n} \times \Big(d+ \log (\frac{1}{\delta})\Big) \right) +   \mathcal{O}\left( \frac{1}{T^{\frac{2\alpha-2}{3\alpha-2}}} \log\frac{1}{\delta} \right).
\end{align*}

Taking $T \asymp (\frac{n}{d})^{\frac{3\alpha-2}{5\alpha-4}}$, we have the following inequality with probability $1 - 2\delta$
 \begin{align*}
&\frac{1}{T} \sum_{t=1}^T \| \nabla F(\mathbf{w}_{t}) \|^2 \\& = 
            \mathcal{O} \left( \Big(\frac{d}{n}\Big)^{\frac{2\alpha-2}{5\alpha-4}} \log(1/\delta) \log \left(1+(\frac{n}{d})^{\frac{3\alpha}{5\alpha-4}}\right)\right),
\end{align*}
which means 
with probability at least $1 - \delta$ we have
 \begin{align*}
&\frac{1}{T} \sum_{t=1}^T \| \nabla F(\mathbf{w}_{t}) \|^2 \\& = 
            \mathcal{O} \left( \Big(\frac{d}{n}\Big)^{\frac{2\alpha-2}{5\alpha-4}} \log(1/\delta) \log \left(1+(\frac{n}{d})^{\frac{3\alpha}{5\alpha-4}}\right)\right).
\end{align*}
The proof is complete.
\end{proof}
\subsection{Proof of Theorem 17}\label{appendib77}
\begin{proof}
 With the descent lemma of smoothness, we have
\begin{align*}
&F_S(\mathbf{w}_{t+1}) - F_S(\mathbf{w}_{t}) \\&\leq \langle \mathbf{w}_{t+1} - \mathbf{w}_{t}, \nabla F_S(\mathbf{w}_{t}) \rangle + \frac{1}{2} L \| \mathbf{w}_{t+1} -\mathbf{w}_{t}  \|^2\\
& = -\eta_t \langle  \nabla \bar{f}(\bar{\mathbf{w}}_t;z_{j_t}), \nabla F_S(\mathbf{w}_{t}) \rangle + \frac{1}{2} L \eta_t^2 \|\nabla \bar{f}(\bar{\mathbf{w}}_t;z_{j_t}) \|^2 \\
& = -\eta_t \langle  \nabla \bar{f}(\bar{\mathbf{w}}_t;z_{j_t}) - \nabla F_S(\bar{\mathbf{w}}_t) + \nabla F_S(\bar{\mathbf{w}}_t), \\&\nabla F_S(\mathbf{w}_{t}) -   \nabla F_S(\bar{\mathbf{w}}_t) +    \nabla F_S(\bar{\mathbf{w}}_t) \rangle + \frac{1}{2} L \eta_t^2 \|\nabla \bar{f}(\bar{\mathbf{w}}_t;z_{j_t}) \|^2 \\
& = -\eta_t \langle  \nabla \bar{f}(\bar{\mathbf{w}}_t;z_{j_t}) , \nabla F_S(\mathbf{w}_{t}) -   \nabla F_S(\bar{\mathbf{w}}_t) \rangle \\&-\eta_t   \langle  \nabla \bar{f}(\bar{\mathbf{w}}_t;z_{j_t}) - \nabla F_S(\bar{\mathbf{w}}_t) , \nabla F_S(\bar{\mathbf{w}}_t) \rangle\\
& -\eta_t \| \nabla F_S(\bar{\mathbf{w}}_{t}) \|^2 + \frac{1}{2} L \eta_t^2 \|\nabla \bar{f}(\bar{\mathbf{w}}_t;z_{j_t}) \|^2 \\
&\leq \frac{L}{2}\eta_t^2 \| \nabla \bar{f}(\bar{\mathbf{w}}_t;z_{j_t}) \|^2 + \frac{L}{2} \| \mathbf{w}_{t} -  \bar{\mathbf{w}}_t \|^2 \\&-\eta_t   \langle  \nabla \bar{f}(\bar{\mathbf{w}}_t;z_{j_t}) - \nabla F_S(\bar{\mathbf{w}}_t) , \nabla F_S(\bar{\mathbf{w}}_t) \rangle\\
& -\eta_t \| \nabla F_S(\bar{\mathbf{w}}_{t}) \|^2 + \frac{1}{2} L \eta_t^2 \|\nabla \bar{f}(\bar{\mathbf{w}}_t;z_{j_t}) \|^2,
\end{align*}
where the last inequality follows from Young's inequality and $\|  \nabla F_S(\mathbf{w}_{t}) -   \nabla F_S(\bar{\mathbf{w}}_t)  \| \leq L \|\mathbf{w}_{t} -  \bar{\mathbf{w}}_t \|$.
The above inequality implies that 
\begin{align*}
&\| \nabla F_S(\bar{\mathbf{w}}_{t}) \|^2 \leq \frac{ F_S(\mathbf{w}_{t}) - F_S(\mathbf{w}_{t+1}) }{\eta_t}  \\&-  \langle  \nabla \bar{f}(\bar{\mathbf{w}}_t;z_{j_t}) - \nabla F_S(\bar{\mathbf{w}}_t) , \nabla F_S(\bar{\mathbf{w}}_t) \rangle  
\\&+  L \eta_t  \|\nabla \bar{f}(\bar{\mathbf{w}}_t;z_{j_t}) \|^2 + \frac{L}{2\eta_t} \| \mathbf{w}_{t} -  \bar{\mathbf{w}}_t \|^2.
\end{align*}
Following the proof technique in (\ref{thelastinequali}), we can get
\begin{align*}
&\sum_{t=1}^T \| \nabla F_S(\bar{\mathbf{w}}_{t}) \|^2 
\leq  (2M+2L)\sqrt{G_0^2 + \sum_{t=1}^T \|\nabla \bar{f}(\bar{\mathbf{w}}_t;z_{j_t})\|^2} \\& -\sum_{t=1}^T \langle  \nabla \bar{f}(\bar{\mathbf{w}}_t;z_{j_t}) - \nabla F_S(\bar{\mathbf{w}}_t) , \nabla F_S(\bar{\mathbf{w}}_t) \rangle\\& + \sum_{t=1}^T \frac{L}{2\eta_t} \| \mathbf{w}_{t} -  \bar{\mathbf{w}}_t \|^2.
\end{align*}
According to (\ref{danxinjiayi}) and (\ref{youdiandanxin}), with probability at least $1-\delta$, we get
\begin{align}\label{eq676}\nonumber
&\frac{1}{4}\sum_{t=1}^T \| \nabla F_S(\bar{\mathbf{w}}_{t}) \|^2\leq  (2M+2 L)\sqrt{G_0^2 + \sum_{t=1}^T \|\nabla \bar{f}(\bar{\mathbf{w}}_t;z_{j_t})\|^2} \\\nonumber
&  + \frac{1}{2}G^{2\alpha} \tau^{2-2\alpha}T +   \max\{4  \tau^{2-\alpha} G^{\alpha}, 2  \tau G\} \log(1/\delta) \\
&+ \sum_{t=1}^T \frac{L}{2\eta_t} \| \mathbf{w}_{t} -  \bar{\mathbf{w}}_t \|^2.
\end{align}
To bound the term $\sum_{t=1}^T \frac{L}{2\eta_t} \| \mathbf{w}_{t} -  \bar{\mathbf{w}}_t \|^2$, we introduce the following lemma.
\begin{lem}[Proposition 5.2 in \cite{kavis2021high}]\label{lemma7}
When $\eta_t = \beta_t$, there holds
\begin{align*}
\sum_{t=1}^T \frac{1}{\eta_t} \| \mathbf{w}_{t} -  \bar{\mathbf{w}}_t \|^2 = 0
\end{align*}
When $|\eta_t - \beta_t| = \alpha_t \lambda_t$, if  $\alpha_t =\frac{2}{t+1}$, there holds 
\begin{align*}
&\sum_{t=1}^T \frac{1}{\eta_t} \| \mathbf{w}_{t} -  \bar{\mathbf{w}}_t \|^2  \\&\leq  
2 \sqrt{G_0^2 + \sum_{t=1}^T \|\nabla \bar{f}(\bar{\mathbf{w}}_t;z_{j_t})\|^2} \sum_{t=1}^T \lambda_t^2 \|\nabla f(\bar{\mathbf{w}}_t;z_{j_t})\|^2.
\end{align*}
\end{lem}
Thus, by Lemma \ref{lemma7} and Lemma 22, we get if  $ \alpha_t =\frac{2}{t+1}$
\begin{align*}
&\sum_{t=1}^T \frac{L}{2\eta_t} \| \mathbf{w}_{t} -  \bar{\mathbf{w}}_t \|^2 \\&\leq  
L \sqrt{G_0^2 + \sum_{t=1}^T \|\nabla \bar{f}(\bar{\mathbf{w}}_t;z_{j_t})\|^2}\Big(1+\log (1+\sum_{t=1}^T \|\nabla f(\bar{\mathbf{w}}_t;z_{j_t})\|^2) \Big).
\end{align*}


Due to $\tau  = p T^{\frac{1}{3\alpha - 2}}$, the above inequality implies that if  $\alpha_t =\frac{2}{t+1}$
\begin{align*}
&\sum_{t=1}^T \frac{L}{2\eta_t} \| \mathbf{w}_{t} -  \bar{\mathbf{w}}_t \|^2 \\&\leq  
L \sqrt{G_0^2 + \sum_{t=1}^T \|\nabla \bar{f}(\bar{\mathbf{w}}_t;z_{j_t})\|^2}\left(1+\log (1+p T^{\frac{3\alpha}{3\alpha - 2}}) \right).
\end{align*}


Next, we bound the term $\sum_{t=1}^T \|\nabla \bar{f}(\bar{\mathbf{w}}_t;z_{j_t})\|^2$. Following the proof method of (\ref{eq007}) in the Proof of Theorem 13, with probability at least $1-\delta$, when $\tau \leq  T^{\frac{1}{\alpha}}$, i.e., $p\leq T^{\frac{2\alpha-2}{\alpha(3\alpha-2)}}$, we can get
\begin{align*}
&\sum_{t=1}^T \|\nabla \bar{f}(\bar{\mathbf{w}}_t;z_{j_t})\|^2 \\&
\leq \mathcal{O}(TT^{\frac{2-\alpha}{3\alpha-2}}\log(1/\delta)) +3\sum_{t=1}^T \| \nabla F_S(\bar{\mathbf{w}}_t) \|^2.
\end{align*}
Plugging these bounds into (\ref{eq676}), we get the following inequality with probability $1-2\delta$
\begin{align}\label{eq679}\nonumber
&\frac{1}{4}\sum_{t=1}^T \| \nabla F_S(\bar{\mathbf{w}}_{t}) \|^2\leq  (2M+2 L)\\\nonumber&\times\sqrt{G_0^2 + \mathcal{O}(TT^{\frac{2-\alpha}{3\alpha-2}}\log(1/\delta)) +3\sum_{t=1}^T \| \nabla F_S(\bar{\mathbf{w}}_t) \|^2} \\\nonumber
&  + \frac{1}{2}G^{2\alpha} \tau^{2-2\alpha}T +   \max\{4  \tau^{2-\alpha} G^{\alpha}, 2  \tau G\} \log(1/\delta) \\&+ \sum_{t=1}^T \frac{L}{2\eta_t} \| \mathbf{w}_{t} -  \bar{\mathbf{w}}_t \|^2.
\end{align}
Solving the quadratic inequality of $\sum_{t=1}^T \| \nabla F_S(\bar{\mathbf{w}}_t) \|^2$, (\ref{eq679}) implies that with probability $1-2\delta$
\begin{align*}
&\sum_{t=1}^T \| \nabla F_S(\bar{\mathbf{w}}_t) \|^2 \leq  \mathcal{O}\left(\frac{1}{T^\frac{\alpha}{3\alpha - 2}}\log \frac{1}{\delta}\right).
\end{align*}
We finally obtain the following inequality with probability $1-\delta$
\begin{align*}
&\frac{1}{T}\sum_{t=1}^T \| \nabla F_S(\bar{\mathbf{w}}_{t}) \|^2
\leq \mathcal{O}\left(\frac{\log 1/\delta}{T^\frac{2\alpha -2}{3\alpha - 2}}\right).
\end{align*}
The proof is complete.
\end{proof}
\end{document}